\documentclass{article} 
\usepackage{iclr2027_conference,times}

\usepackage{amsmath,amsfonts,bm}

\def\eqref#1{equation~\ref{#1}}

\def\1{\bm{1}}

\DeclareMathAlphabet{\mathsfit}{\encodingdefault}{\sfdefault}{m}{sl}
\SetMathAlphabet{\mathsfit}{bold}{\encodingdefault}{\sfdefault}{bx}{n}

\usepackage{hyperref}
\usepackage{url}
\usepackage[utf8]{inputenc} 
\usepackage[T1]{fontenc}    
\usepackage{hyperref}       
\usepackage{url}            
\usepackage{booktabs}       
\usepackage{amsfonts}       
\usepackage{nicefrac}       
\usepackage{microtype}      
\usepackage{xcolor}         
\usepackage{lipsum}
\usepackage{enumitem}
\usepackage{amsmath}
\usepackage{graphicx}
\usepackage{multirow}
\usepackage[most]{tcolorbox}
\usepackage{xspace}
\usepackage{fontawesome5}
\usepackage{caption}

\usepackage[table]{xcolor}

\setlipsum{
  par-before = \begingroup\color{gray},
  par-after = \endgroup
}

\newcommand{\ourmethod}{\textsc{DAR}\xspace}

\title{Reinforcement Learning over Predictive \\ Distributions for LLM Regression}

\author{%
  \hspace*{-0.4cm}%
  \makebox[\linewidth][c]{%
    \bfseries Jungsoo Park$^{1}$,
    Hyungjoo Chae$^{1}$,
    Ethan Mendes$^{1}$,
    Jay DeYoung$^{2}$%
  }\\
  \hspace*{-0.4cm}%
  \makebox[\linewidth][c]{%
    \bfseries Varsha Kishore$^{2}$,
    Wei Xu$^{1}$,
    Alan Ritter$^{1}$%
  }\\
  \hspace*{-0.4cm}%
  \makebox[\linewidth][c]{\normalfont $^{1}$Georgia Institute of Technology}\\
  \hspace*{-0.4cm}%
  \makebox[\linewidth][c]{\normalfont $^{2}$Allen Institute for AI}
}

\iclrfinalcopy 
\begin{document}

\maketitle

\begin{abstract}

Large language models (LLMs) have emerged as flexible regressors capable of predicting real-valued quantities from heterogeneous inputs. 
Yet most LLM regression objectives optimize predictions independently, often yielding poor calibration. 
We introduce \emph{Distribution-Aware Reward} (DAR), an on-policy reinforcement learning objective that instead jointly evaluates the empirical predictive distribution formed by multiple predictions for the same input. 
To translate this distribution-level objective into rollout-level rewards, we assign each prediction credit based on its leave-one-out contribution to the quality of the overall predictive distribution. 
This encourages predictions that are well-centered and appropriately dispersed around the target.
We evaluate on three regression settings: a synthetic task probing interpolation and extrapolation, and two real-world scientific tasks involving code and molecular data. 
Across tasks, DAR produces better-calibrated uncertainty estimates while consistently reducing prediction error and improving ranking quality over supervised fine-tuning and pointwise reinforcement learning.
Together, these results highlight the benefits of distribution-aware training for LLM regression.\footnote{\href{https://jjumssu.github.io/dar-project-page/}
{\faGithub\ Code \& Data}.}

\end{abstract}

\section{Introduction}
\label{sec:intro}

Many prediction problems in science and engineering require estimating real-valued quantities from heterogeneous inputs, including experimental protocols, source code, and molecular representations~\citep{wu2018moleculenet, park2025precog, akhauri2025regression}. 
Large language models (LLMs) provide a flexible approach to such problems when inputs can be represented as sequences~\citep{requeima2024llm, song2024omnipred, tang2024understanding}. 
Recent work has improved LLM regression through in-context learning~\citep{vacareanu2024words}, supervised fine-tuning (SFT)~\citep{song2024omnipred, lukasik2025better}, and reinforcement learning (RL) with numeric rewards~\citep{chen2025beyond}.

\begin{figure}[t]
    \centering
    \includegraphics[width=1.0\columnwidth]{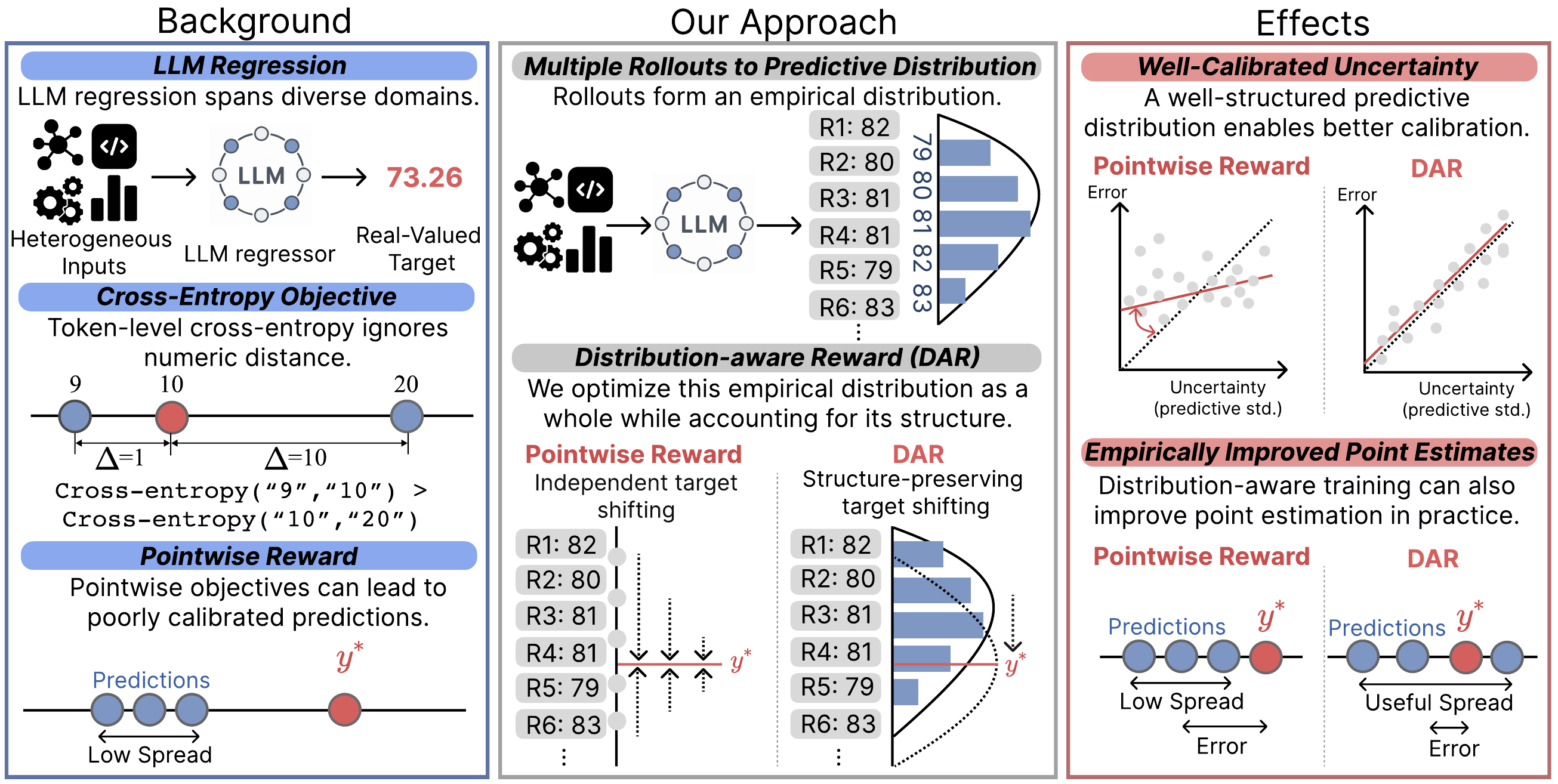}
        \caption{\textbf{Distribution-Aware Reward.}
    \textbf{Left:} Standard objectives optimize individual predictions without explicitly modeling their distribution.
    \textbf{Middle:} Multiple LLM rollouts form an empirical predictive distribution. Pointwise rewards score these rollouts independently, whereas \ourmethod rewards each rollout according to its contribution to the distribution as a whole, encouraging accurate predictions while preserving useful spread.
    \textbf{Right:} Distribution-aware training produces better-calibrated predictive distributions, while empirically also yielding stronger aggregate point estimates.}
    \label{fig:main_figure}
\end{figure}

Beyond producing a single prediction, repeated sampling from an LLM naturally induces a predictive distribution over possible values. 
This distribution can be aggregated for point prediction while also providing information about predictive uncertainty.
Yet existing training objectives largely ignore this distributional structure. 
They typically score each decoded prediction independently according to its distance from the target~\citep{chen2025beyond}, treating rollouts as separate predictions rather than samples from a shared predictive distribution (Figure~\ref{fig:main_figure} (left)). 
We refer to these as \emph{pointwise objectives}. 
While effective for encouraging individual predictions toward the target, pointwise squared-error rewards also penalize variation across samples, which can suppress predictive variation even when uncertainty is warranted.
Ideally, an LLM regressor should produce both accurate aggregate predictions and predictive distributions that support calibrated uncertainty estimates~\citep{gneiting2007probabilistic, gneiting2007strictly, levi2022evaluating}.

We therefore introduce \emph{Distribution-Aware Reward} (\ourmethod), an RL objective that optimizes the predictive distribution formed by multiple decoded samples. 
The key idea is to shift the unit of optimization from individual predictions to the distribution they collectively form. 
As illustrated in Figure~\ref{fig:main_figure} (middle), rather than rewarding each rollout solely for being close to the target, \ourmethod assigns credit based on how each rollout contributes to the quality of the predictive distribution as a whole.

\ourmethod treats multiple rollouts as an empirical predictive distribution rather than scoring each one in isolation.
We measure the quality of this distribution using the Continuous Ranked Probability Score (CRPS), a scoring rule that rewards distributions that place probability mass near the target while maintaining appropriate spread~\citep{matheson1976scoring, gneiting2007strictly}.
Thus, \ourmethod does not encourage dispersion for its own sake, but balances proximity to the target with dispersion across predictions.
We then use leave-one-out credit assignment, rewarding each rollout according to how much its inclusion improves the overall distribution. 
This provides a joint training signal that accounts for interactions among sampled predictions.

We evaluate \ourmethod on three regression settings: a synthetic task with multimodal target distributions~\citep{fishkov2025uncertainty}, code performance prediction~\citep{akhauri2025regression}, and molecular property prediction~\citep{wu2018moleculenet}.
Across these tasks, \ourmethod produces predictive distributions with improved calibration and uncertainty estimates that better align with prediction error.
We further analyze the learned predictive distributions and rollout behavior during training to characterize how distribution-aware optimization changes the model’s sampling behavior.

Importantly, these distributional benefits do not come at the expense of predictive performance.
\ourmethod consistently reduces prediction error and improves rank correlation over supervised fine-tuning and pointwise RL baselines.
The gains are particularly strong for code performance prediction, where \ourmethod improves Spearman correlation by up to 6 points.
It also achieves the strongest rank correlation on the synthetic task and remains competitive with specialized graph and 3D molecular models for molecular property prediction, despite using only SMILES, a text-based molecular representation.
These improvements come with negligible computational overhead over standard RL training.
Together, our results show that optimizing LLM predictions at the distribution level can support calibrated uncertainty while also improving aggregate point prediction and ranking.

\section{DAR: Distribution-Aware Reward for LLM Regression}
\label{sec:method}

We introduce \ourmethod, a distribution-aware reward for on-policy RL that evaluates multiple predictions for the same input jointly rather than independently. 
\ourmethod can be used within standard rollout-based RL frameworks such as GRPO~\citep{shao2024deepseekmath} by replacing the pointwise reward with a distribution-aware reward assigned to each rollout. 
We first review standard training objectives for LLM regression and then derive the distribution-level objective and its rollout-level credit assignment.

\subsection{Background: Training Large Language Models for Regression}
\label{sec:method-background}
LLMs have recently emerged as flexible regressors capable of handling heterogeneous inputs.
In this setting, given an input $x$ and target $y$, an autoregressive language model $\pi_\theta$ generates a sequence of output tokens $z$.
For regression, this output is mapped to a scalar prediction $p \in \mathbb{R}$ using a deterministic parsing function $g$, such that $p=g(z)$.
A common approach to training LLMs for regression is supervised fine-tuning (SFT)~\citep{song2024omnipred, akhauri2025regression, lukasik2025better}.
Given an input–target pair $(x,y)$, we represent the numeric target $y$ as a sequence $z^\star$, corresponding to its string representation.
SFT then minimizes the negative log-likelihood of this target sequence:
\setlength{\abovedisplayskip}{10pt}
\setlength{\belowdisplayskip}{10pt}
\begin{align}
\mathcal{L}_{\mathrm{SFT}}(\theta)
=
-\log \pi_\theta(z^\star \mid x).
\end{align}
Although effective, this objective optimizes token-level prediction of the target representation rather than the quality of the resulting numeric prediction. One alternative is to formulate regression as classification over a discretized set of numeric values~\citep{lukasik2025better}.
The model predicts a probability distribution over these values, which is then aggregated to produce a continuous prediction. While this formulation retains standard cross-entropy training, it requires the numeric output space to be discretized in advance, which can be restrictive when target ranges vary substantially across regression tasks. 

Reinforcement learning (RL) provides another alternative by defining rewards directly on decoded numeric predictions.
Unlike SFT, RL can sample multiple outputs for the same input and assign rewards based on their resulting numeric values.
For the same input-target pair $(x,y)$, we sample $K$ textual rollouts from the current policy,
\begin{align}
z_k \sim \pi_\theta(\cdot \mid x), &\qquad k=1,\ldots,K, \\
p_k &= g(z_k),
\end{align}
where $p_k\in\mathbb{R}$ is the numeric prediction extracted from the $k$-th rollout. This yields a set of predictions $\{p_k\}_{k=1}^K$ for the same target $y$. A standard pointwise reward evaluates each rollout independently. For example, a mean squared error (MSE)-based reward~\citep{koa2025reasoning, chen2025beyond} can be expressed as:
\begin{align}
R^{\mathrm{MSE}}_k
=
-(p_k-y)^2.
\end{align}
This directly optimizes the decoded numeric output rather than its token-level representation, but evaluates each rollout independently of the other predictions sampled for the same input. As a result, pointwise rewards do not account for the distribution collectively formed by multiple rollouts. Our goal is not to encourage greater dispersion for its own sake, but to optimize the quality of this predictive distribution by accounting for both accuracy and uncertainty. These considerations motivate our design of \ourmethod.

\subsection{Rewarding Predictive Distribution Quality}
\label{sec:method-dar}

To move beyond pointwise rewards, we propose a distribution-aware reward (\ourmethod) that jointly evaluates the $K$ rollouts as an empirical predictive distribution rather than rewarding each prediction in isolation. We seek an objective that captures both proximity to the target and the dispersion of the sampled predictions, without imposing a parametric assumption on their distribution. The Continuous Ranked Probability Score (CRPS)~\citep{matheson1976scoring,gneiting2007strictly} satisfies these properties as a proper scoring rule that evaluates a predictive distribution against a realized numerical outcome, with its expected score optimized when the predictive distribution matches the true outcome distribution. Since CRPS is conventionally minimized, we define the distribution-level reward over the entire rollout set $\{p_k\}_{k=1}^K$ using its negative empirical form:
\begin{align}
\label{eq:dar-crps}
R^{\mathrm{dist}}
=
\underbrace{
-\frac{1}{K}\sum_{k=1}^{K}\lvert p_k-y\rvert
}_{\text{distance to target}}
+
\underbrace{
\frac{1}{2K^2}\sum_{k=1}^{K}\sum_{\ell=1}^{K}
\lvert p_k-p_\ell\rvert
}_{\text{spread among predictions}}.
\end{align}
The first term is the negative mean absolute error (MAE) of the sampled predictions, rewarding proximity to the target. By itself, it encourages every rollout to move toward $y$, irrespective of the other $K-1$ predictions, suppressing variation across the sampled distribution.
The second term counterbalances this tendency by rewarding pairwise separation among the sampled predictions. Importantly, the second term does not by itself encourage predictions to concentrate around the target. Rather, the two terms jointly trade off proximity to the observed target against dispersion among predictions, discouraging both excessive concentration and excessive spread.


All $K$ predictions for the same input $x$ jointly determine $R^{\mathrm{dist}}$. To use this distribution-level objective for on-policy RL optimization, we need to assign a scalar reward to each rollout. The key challenge is thus to attribute the quality of the sampled distribution back to the individual prediction that constitutes it. We address this credit-assignment problem using a leave-one-out marginal contribution. Let $R^{\mathrm{dist}}_{-k}$ denote the distribution-level reward computed after removing the $k$th rollout $p_k$ from the sampled set. We define the rollout-level \ourmethod reward as follows, using $R^{\mathrm{dist}}$ and $R^{\mathrm{dist}}_{-k}$, both computed over sets of sampled rollouts:
\begin{align}
\label{eq:dar-loo}
R^{\mathrm{DAR}}_k
=
R^{\mathrm{dist}}
-
R^{\mathrm{dist}}_{-k}.
\end{align}
Intuitively, if removing $p_k$ lowers the distribution-level reward, then $R^{\mathrm{DAR}}_k>0$, indicating that the rollout improves the overall predictive distribution.
If removing it increases the reward, then $R^{\mathrm{DAR}}_k<0$, indicating that this particular rollout degrades the predictive distribution.
Although this leave-one-out construction makes each rollout’s reward depend on the other sampled predictions, it introduces no additional bias into the policy-gradient estimator; we provide a formal analysis in Appendix~\ref{sec:loo_crps_properties}.

\paragraph{Motivating Example: MSE vs.\ DAR.}

We illustrate that DAR can upweight a rollout with larger pointwise error when it improves the \emph{collective} predictive distribution.
Consider a single input with target $y=5$ and three rollouts producing predictions $[3,\,4,\,8]$.
A pointwise MSE-based reward scores rollouts independently and therefore favors $4$ (lowest error) and $3$, while penalizing $8$, encouraging concentration around near-target values.

DAR instead scores the \emph{set} of rollouts via CRPS and assigns credit by marginal contribution.
Here, $3$ and $4$ lie on the same side of the target, while $8$ provides a prediction on the other side.
Although $8$ has larger pointwise error, removing it worsens CRPS more than removing $3$.
DAR therefore assigns higher marginal credit to $8$ than to the more redundant prediction $3$.

\section{Experimental Setup}
\label{sec:exp}

We evaluate whether \ourmethod improves predictive performance and uncertainty estimation across diverse regression settings. 
Our evaluation spans three regression settings with different input modalities and prediction challenges: a synthetic regression task designed to probe interpolation and extrapolation, code performance prediction, and molecular property prediction.

\subsection{Tasks}
\label{sec:exp-tasks}

We select three regression settings that span distinct input representations and prediction challenges, including both controlled and established real-world benchmarks.
Together, they allow us to evaluate \ourmethod in a controlled distributional setting as well as on real-world scientific regression problems involving code and molecular representations.
Further details are provided in Appendix~\ref{sec:appendix_experimental_setup}.

\paragraph{Synthetic Distributional Regression.}

We use the synthetic regression benchmark introduced by~\citet{fishkov2025uncertainty}, where the target is sampled from a known heteroscedastic two-component Gaussian mixture conditioned on a scalar input $x$. 
This controlled setting provides access to the ground-truth conditional distribution, making it useful for evaluating both predictive accuracy and uncertainty behavior under interpolation and extrapolation. 
We train on 1,200 samples from the ground-truth process and evaluate on disjoint test regions with 600 samples covering both interpolation and extrapolation. 
We report RMSE and Spearman correlation.

\paragraph{Coding Performance Prediction.}

We evaluate on two code performance prediction benchmarks introduced by~\citet{akhauri2025regression}: Triton kernel latency prediction (KBSS) and Python peak-memory prediction (APPS). These tasks require predicting continuous execution properties directly from source code and provide established benchmarks for evaluating LLM-based regression on complex sequential inputs. We use the dataset splits from prior work, with 6K/1K/1K examples for training, validation, and testing, and report Spearman correlation and MAE.

\paragraph{Molecular Property Prediction.}

We evaluate molecular property prediction on three established MoleculeNet benchmarks~\citep{wu2018moleculenet}: ESOL, FreeSolv, and Lipophilicity. Each benchmark defines a separate regression task over molecules represented as SMILES strings~\citep{weininger1988smiles}: ESOL predicts aqueous solubility, FreeSolv predicts hydration free energy, and Lipophilicity predicts a logD-based measure of hydrophobicity. These datasets are widely used benchmarks for molecular property prediction and allow us to compare LLM-based regression against established molecular models that use graph or 3D representations. We use the standard scaffold splits following~\citet{zhou2023unimol} and report RMSE.

\subsection{Implementation Details}
\label{sec:exp-impl}

We follow a standard on-policy RL fine-tuning recipe for sequence models~\citep{schulman2017proximal,ouyang2022training}. Unless otherwise stated, we use a learning rate of $1\times10^{-6}$, train for up to 300 steps, and sample $K=12$ rollouts per input with temperature $1.0$ for exploration. All RL and SFT experiments are implemented in \texttt{verl}~\citep{sheng2024hybridflow}, with KL regularization applied in the loss against the fixed reference policy, initialized from the same checkpoint used to start RL training. We select the checkpoint with the best validation performance and report its test results.

For the synthetic experiment, we use Qwen2.5-0.5B-Instruct~\citep{team2024qwen2}; for real regression tasks, we use Qwen3-4B-Instruct as the default backbone~\citep{yang2025qwen3}. We disable explicit reasoning by default for efficiency and analyze its effect in Appendix~\ref{sec:analysis-reasoningvsnon_reasoning}. During training, each rollout is parsed into a scalar prediction using a strict output format. Invalid rollouts receive method-specific penalties, such as the minimum batch reward for the negative-MSE baseline. At evaluation time, we generate 32 samples and average them as the final prediction, following standard LLM regression practice~\citep{lukasik2024rail, lukasik2025better, akhauri2025regression}. We repeat evaluation with 5 random seeds and report averaged metrics~\citep{chen2025beyond}.

\subsection{Baselines}
\label{sec:exp-baselines}

We compare against baselines that cover the main ways LLMs can be used for regression. At the prompting level, we evaluate both proprietary frontier LLMs and the open-weight base model in \emph{ICL} settings~\citep{vacareanu2024words}, capturing off-the-shelf inference without task-specific training. We also evaluate domain-specific LLMs, such as ChemLLM~\citep{zhang2024chemllm, yu2024llasmol}, in the zero-shot setting.

At the training level, we consider a value-head regression baseline trained with MSE~\citep{tang2024understanding} and an SFT baseline that fine-tunes the model to generate a single numeric output~\citep{lukasik2025better}. Finally, in the on-policy RL setting, we compare conventional pointwise reward optimization based on negative MSE~\citep{chen2025beyond} against DAR, which directly optimizes the predictive distribution induced by multiple rollouts.

\begin{figure}[th]
    \centering
    \includegraphics[width=0.9\columnwidth]{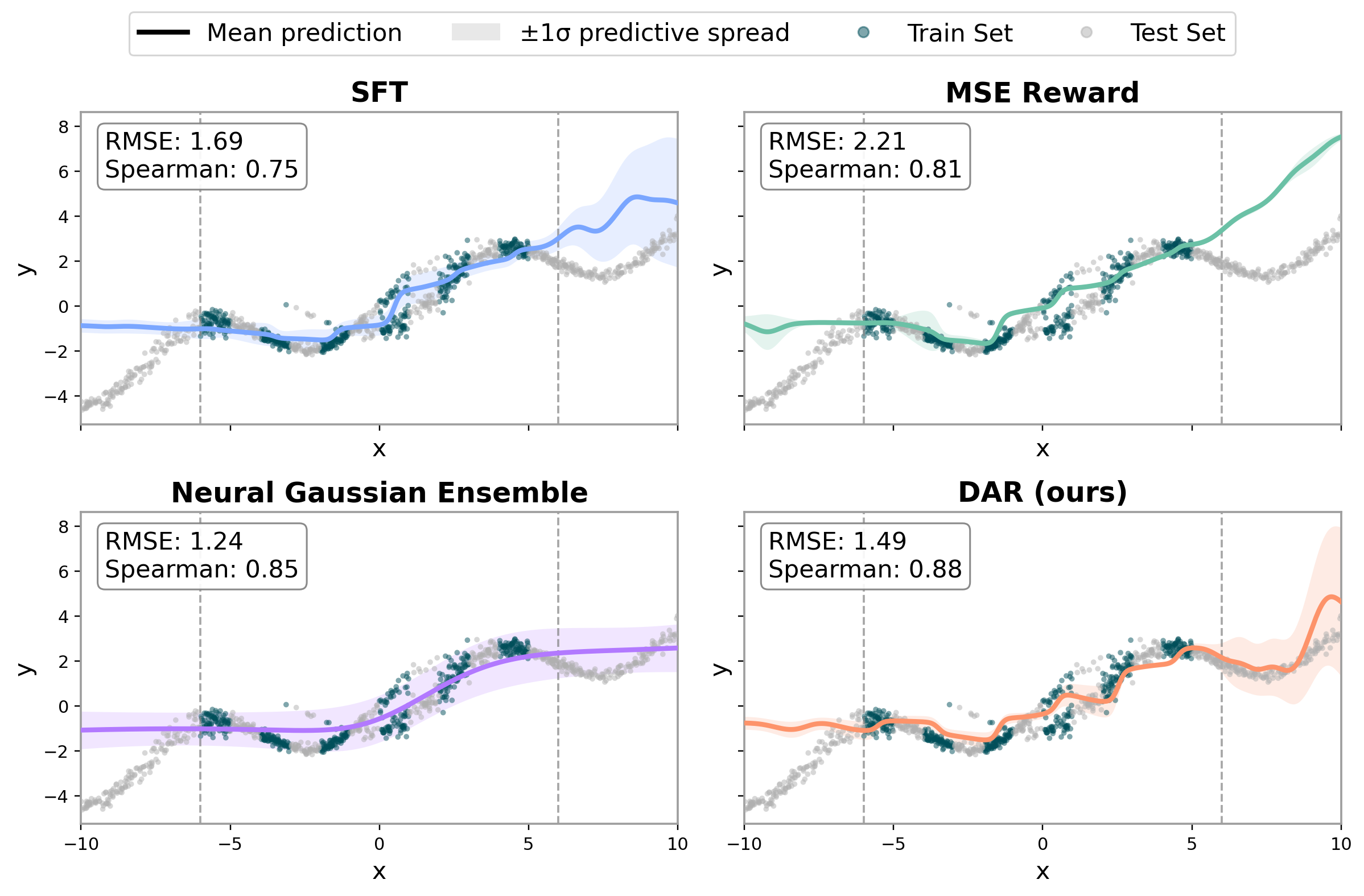}
    \caption{
    \textbf{Synthetic distributional regression.}
    Mean predictions and $\pm 1\sigma$ predictive spread on the 1D Gaussian mixture task.
    Dashed vertical lines at $x=\pm 6$ mark the boundary between interpolation and extrapolation regions.
    }
    \label{fig:synthetic_1d_representation}
\end{figure}

\section{Experimental Results}
\label{sec:experimental_results}

We evaluate \ourmethod on point estimation and ranking across the three regression settings. 
We analyze uncertainty estimation and calibration separately in Section~\ref{sec:analysis-calibration}.

\subsection{Synthetic Distributional Regression}
\label{sec:experimental_results-synthetic_regression}

Figure~\ref{fig:synthetic_1d_representation} compares SFT, MSE reward shaping, a Neural Gaussian Ensemble~\citep{fishkov2025uncertainty}, and \ourmethod on the synthetic 1D Gaussian mixture task. \ourmethod achieves the highest Spearman correlation among the LLM-based methods and better follows the target structure, particularly in the extrapolation region $(6,10)$, while SFT and MSE show larger deviations. 

The Neural Gaussian Ensemble achieves lower RMSE, as expected from its task-matched Gaussian inductive bias. In contrast, \ourmethod achieves strong rank correlation without task-specific inductive biases or parametric assumptions. This controlled experiment therefore illustrates how distribution-aware optimization can improve learned prediction structure rather than demonstrating specialization to Gaussian mixtures.

\subsection{Performance Prediction}
\label{sec:experimental_results-performance_prediction}

\begin{table*}[t]
\centering
\footnotesize
\setlength{\tabcolsep}{4pt}
\renewcommand{\arraystretch}{1.0}
\begin{tabular}{clcc cc}
\toprule
\textbf{Model} & \multicolumn{1}{c}{\textbf{Method}}
& \multicolumn{2}{c}{\textbf{KBSS (latency)}} 
& \multicolumn{2}{c}{\textbf{APPS (memory)}} \\
\cmidrule(lr){3-4}\cmidrule(lr){5-6}
& 
& \textbf{$\rho\,(\uparrow)$} & \textbf{MAE\,($\downarrow$)}
& \textbf{$\rho\,(\uparrow)$} & \textbf{MAE\,($\downarrow$)} \\
\midrule
Qwen3-30B-A3B & ICL~\citep{vacareanu2024words}
& 0.32 & 0.027 & 0.38 & 1.10e4 \\
RLM-300M & SFT~\citep{akhauri2025regression}
& 0.59 & 0.013 & 0.89 & 31.0 \\
RLM-4B & SFT~\citep{akhauri2025regression}
& 0.60 & 0.013 & 0.91 & 29.5 \\
\midrule
Qwen3-4B & ICL~\citep{vacareanu2024words}
& 0.23 & 0.019 & 0.27 & 3.46e8 \\

Qwen3-4B & Value Head~\citep{tang2024understanding}
& 0.23 & 0.025 & 0.65 & 306 \\

Qwen3-4B & SFT~\citep{lukasik2025better}
& 0.61
& 0.013
& 0.93
& 29.5 \\

Qwen3-4B & MSE~\citep{chen2025beyond}
& 0.60 
& 0.012 
& 0.92
& 29.8 \\

\rowcolor[HTML]{FEE2E2}
Qwen3-4B & \ourmethod (ours)
& \textbf{0.66}\rlap{$^{\dagger}$} 
& \textbf{0.011}
& \textbf{0.96}\rlap{$^{\dagger}$}
& \textbf{29.1}\rlap{$^{\dagger}$} \\
\bottomrule
\end{tabular}
\caption{
\textbf{Code performance prediction.}
Results on KBSS latency prediction and APPS peak-memory prediction.
MAE is reported in milliseconds for KBSS and in $10^3$ bytes for APPS.
$^{\dagger}$ denotes statistically significant improvement over the strongest baseline under paired bootstrap testing ($p < 0.05$).
}
\label{tab:perf_pred_results_multi}
\end{table*}

Table~\ref{tab:perf_pred_results_multi} reports code-to-performance prediction results, where \ourmethod achieves the strongest overall performance.
In particular, \ourmethod obtains the best rank correlation and competitive or lower MAE, showing that distribution-aware reward optimization improves code performance prediction beyond standard SFT and pointwise RL rewards. 
The gains are especially clear in Spearman correlation, suggesting better sensitivity to input-dependent variation and more reliable ranking of examples.

Off-the-shelf ICL performs poorly, suggesting that general-purpose LLMs struggle to infer precise continuous performance metrics from source code without task-specific adaptation. The value-head baseline also underperforms, indicating that a regression head alone is insufficient for these long-context tasks. SFT substantially improves performance with task-specific supervision, while \ourmethod further improves over SFT by directly optimizing the predictive distribution. 
Pointwise MSE reward shaping provides limited improvement over SFT, whereas \ourmethod achieves stronger performance, particularly in rank correlation.

\subsection{Molecule Property Prediction}
\label{sec:experimental_results-molecule_property}

\begin{table*}[t]
\centering
\footnotesize
\setlength{\tabcolsep}{3pt}
\renewcommand{\arraystretch}{1.0}
\begin{tabular}{clccc}
\toprule
\textbf{Model} & \multicolumn{1}{c}{\textbf{Method}}
& \textbf{ESOL} 
& \textbf{FreeSolv} 
& \textbf{Lipo} \\
\cmidrule(lr){3-3}\cmidrule(lr){4-4}\cmidrule(lr){5-5}
& 
& \textbf{RMSE ($\downarrow$)}
& \textbf{RMSE ($\downarrow$)}
& \textbf{RMSE ($\downarrow$)} \\
\midrule
D-MPNN & GNN~\citep{yang2019analyzing}
& 1.050 & 2.082 & 0.683 \\

AttentiveFP & Graph attention~\citep{xiong2020attentivefp}
& 0.877 & 2.073 & 0.721 \\

Uni-Mol & 3D molecular representation~\citep{zhou2023unimol}
& 0.788 & 1.480 & 0.603 \\
\midrule
Qwen3-30B-A3B & ICL~\citep{vacareanu2024words}
& 1.886 & 4.160 & 1.427 \\

Mol-Instructions & Molecular instruction~\citep{fang2023mol}
& 2.271 & 6.891 & 1.691 \\

ChemLLM & Chemical LLM~\citep{zhang2024chemllm}
& 1.946 & 10.452 & 1.797 \\

LlaSMol & Molecular LLM~\citep{yu2024llasmol}
& 1.150 & 7.150 & 1.010 \\
\midrule
Qwen3-4B & ICL~\citep{vacareanu2024words}
& 2.133 & 4.368 & 1.284 \\

Qwen3-4B & Value Head~\citep{tang2024understanding}
& 0.893 & 3.460 & 0.946 \\

Qwen3-4B & SFT~\citep{lukasik2025better}
& 0.881 & 2.755 & 0.820 \\

Qwen3-4B & MSE~\citep{chen2025beyond}
& 0.878 & 2.561 & 0.822 \\

\rowcolor[HTML]{FEE2E2}
Qwen3-4B & \ourmethod (ours)
& \textbf{0.871}
& \textbf{2.360}\rlap{$^{\dagger}$}
& \textbf{0.776}\rlap{$^{\dagger}$} \\
\bottomrule
\end{tabular}
\caption{
\textbf{Molecular property prediction.}
RMSE on MoleculeNet regression tasks. ESOL is measured in log molar solubility, FreeSolv in kcal/mol hydration free energy, and Lipophilicity in logD. Graph and molecular ML baselines use task-specific molecular representations, while the LLM methods use only SMILES strings. $^{\dagger}$ denotes statistically significant improvement over the strongest LLM baseline under paired bootstrap testing ($p < 0.05$).
}
\label{tab:mol_results}
\end{table*}

Table~\ref{tab:mol_results} reports MoleculeNet property prediction results. Graph-based and 3D molecular models remain strong baselines because they use chemistry-aware representations, including atom-bond graph structure, curated molecular features, or geometric information~\citep{yang2019analyzing, xiong2020attentivefp, zhou2023unimol}. In contrast, the LLM methods operate only on SMILES strings, which provide a textual representation of molecular structure without explicit graph or 3D inductive biases.

Despite this disadvantage, regression fine-tuning substantially improves LLM performance. \ourmethod consistently outperforms other training strategies, with statistically significant gains on FreeSolv and Lipophilicity. While MSE reward shaping sometimes improves over SFT, \ourmethod achieves the strongest overall results, supporting the benefit of distribution-level optimization over pointwise objectives. ICL~\citep{vacareanu2024words} and value-head fine-tuning~\citep{tang2024understanding} remain less effective, likely due to limited exposure to training examples in ICL and the difficulty of learning randomly initialized numeric prediction modules.

\section{Analysis}

We analyze the effects of distribution-aware reward shaping through predictive distribution quality and calibration, rollout behavior during training, and comparisons with alternative objectives.

\subsection{Predictive Distribution Quality and Calibration}
\label{sec:analysis-calibration}

\begin{figure}[t]
    \centering
    \includegraphics[width=0.9\columnwidth]{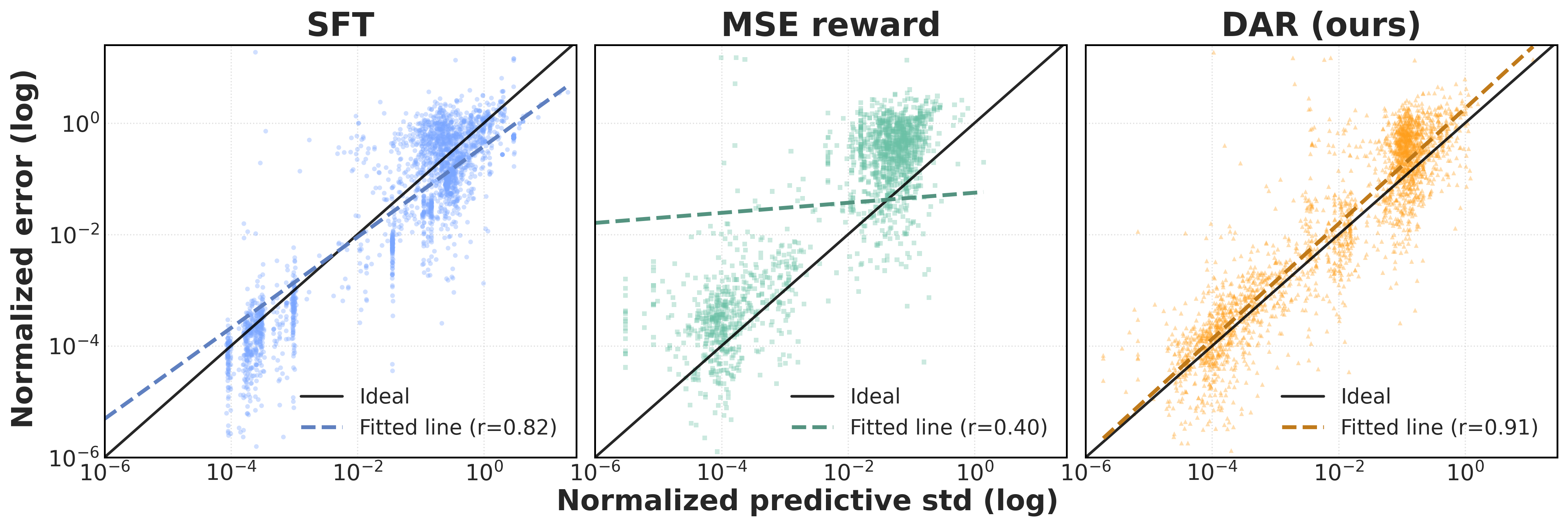}
   \caption{\textbf{Distribution-level calibration and uncertainty diagnostics.} We compare predictive distributions from SFT, MSE, and \ourmethod. The plots show normalized predictive standard deviation versus normalized prediction error on a log-log scale, along with fitted trends and the ideal diagonal.}
    \label{fig:confidence_calibration}
\end{figure}

\paragraph{Rollout Dispersion as Predictive Uncertainty.}
We first examine whether variation across rollouts provides meaningful example-level uncertainty. Following common practice in probabilistic regression~\citep{rasmussen2006gaussian, lakshminarayanan2017deepensembles, kendall2017uncertainties}, we use the standard deviation across sampled predictions as the uncertainty estimate and compare it with the corresponding prediction error. Figure~\ref{fig:confidence_calibration} shows that \ourmethod exhibits the strongest alignment between predictive standard deviation and realized error, suggesting that its rollout dispersion better reflects prediction difficulty than SFT or MSE reward shaping.

\noindent
\begin{minipage}[t]{0.58\columnwidth}
\vspace{0pt}
\paragraph{Predictive Distribution Calibration.}
We next evaluate predictive distribution quality using 90\% prediction-interval coverage error. Coverage error measures calibration by comparing empirical prediction-interval coverage with the nominal level~\citep{kuleshov2018accurate}. In particular, a well-calibrated 90\% prediction interval should contain the true target approximately 90\% of the time, corresponding to a coverage error near zero (the lower the better). We apply the same post-hoc variance recalibration procedure~\citep{laves2020well} to all methods and report macro-averaged performance across the five datasets. \ourmethod reduces the macro-average 90\% coverage error from 0.15 for SFT and 0.42 for MSE to 0.03 (Figure~\ref{fig:interval_calibration}), indicating substantially better agreement with nominal coverage. 
We report additional distributional metrics results in Appendix~\ref{app:crps}.
Together, these results show that distribution-aware training improves both the quality and calibration of the resulting predictive distributions.
\end{minipage}
\hfill
\begin{minipage}[t]{0.39\columnwidth}
\vspace{0pt}
\centering
\includegraphics[width=\linewidth]{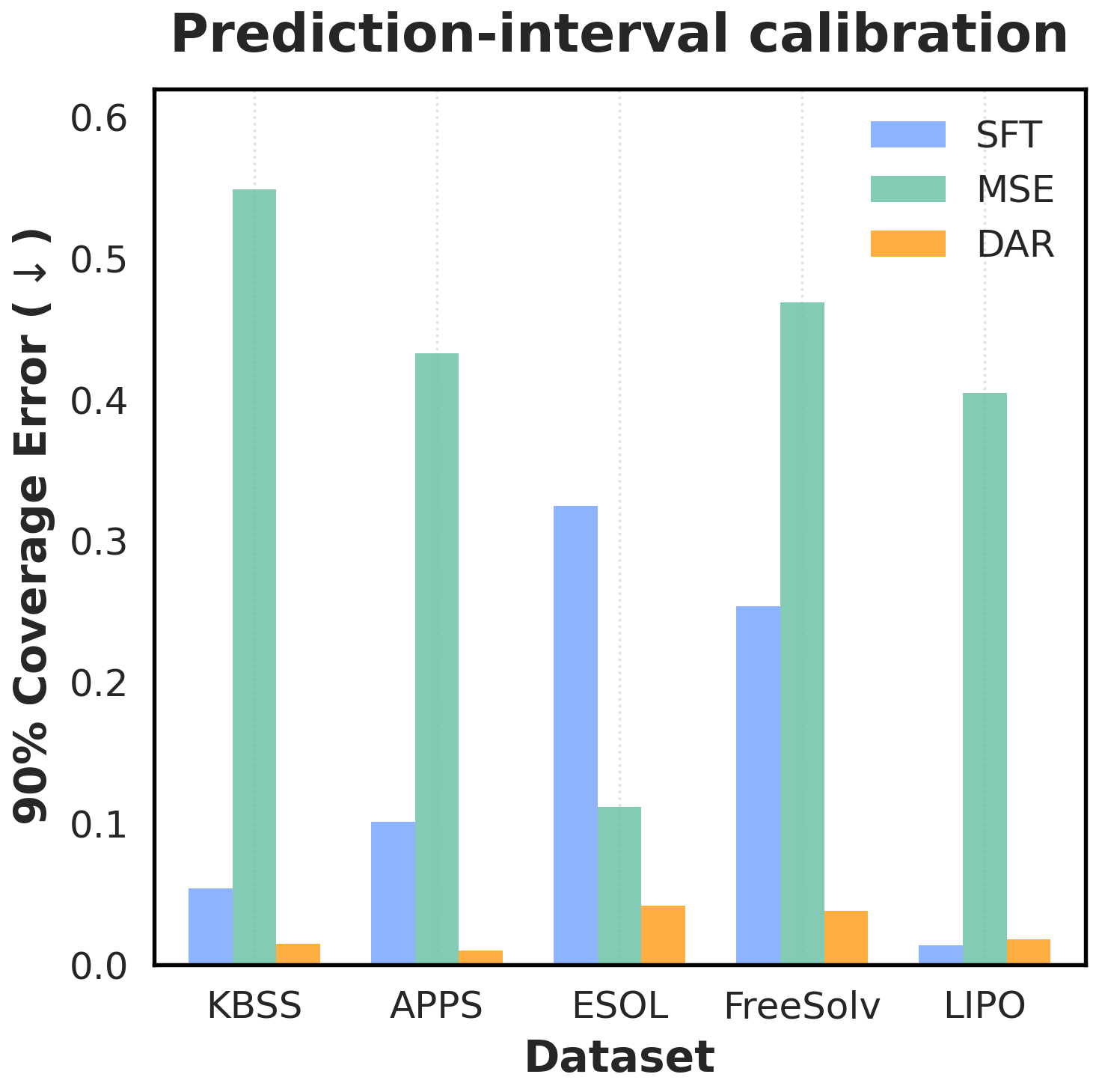}
\captionof{figure}{
\textbf{Prediction-interval calibration.}
90\% coverage error after post-hoc variance recalibration.
}
\label{fig:interval_calibration}
\end{minipage}

\begin{figure}[t]
    \centering
    \includegraphics[width=0.9\columnwidth]{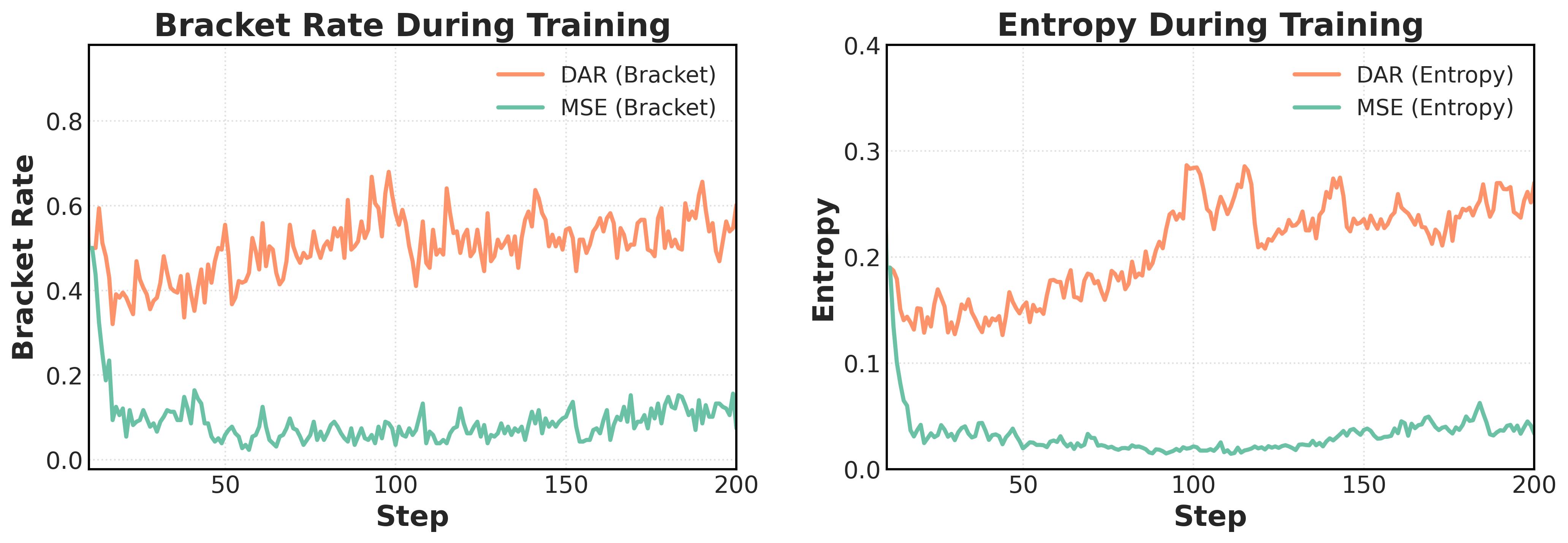}
    \caption{\textbf{Training dynamics during RL under \ourmethod (ours) and MSE.} Bracket rate measures the fraction of examples whose rollout set contains predictions on both sides of the ground truth, while entropy measures prediction diversity, with decreasing entropy indicating potential mode collapse.}
    \label{fig:mae_bracket_comparison}
    \vspace{-.4cm}
\end{figure}

\subsection{Distribution of Rollouts During Training}
\label{sec:analysis-distribution_rollouts}

To examine how distribution-aware reward shaping affects GRPO~\citep{shao2024deepseekmath} training, we analyze rollout diversity, motivated by prior work showing that RL fine-tuning can reduce output diversity or policy entropy~\citep{kirk2023understanding, cui2025entropy}. For each input $x$ with rollout predictions $p_{1:K}$ and target $y$, we measure the bracket rate, $\mathbb{I}\left(\min_k p_{k} \le y \le \max_k p_{k}\right)$, indicating whether the predictions span the ground-truth value. We also track token-level policy entropy during training. Together, these metrics capture target-covering diversity and policy-level stochasticity.

Figure~\ref{fig:mae_bracket_comparison} shows the KBSS results. Compared with MSE reward shaping, \ourmethod achieves a higher bracket rate and maintains higher policy entropy throughout training. 
These results show that distribution-aware reward shaping maintains useful variation across rollouts and improves target coverage, providing a complementary view of how \ourmethod changes sampling behavior during training.

\subsection{Alternative Objectives}
\label{sec:alternative_objectives}

\noindent
\begin{minipage}[t]{0.55\columnwidth}
\vspace{0pt}
We study whether simply increasing diversity or using distributional objectives can reproduce the gains of \ourmethod (Table~\ref{tab:alternative_objectives}). Adding variance regularization slightly improves on LIPO but not KBSS, while entropy regularization provides no consistent benefit. Standard value-head-based distributional objectives, including Gaussian NLL, CRPS, and quantile loss, also underperform MSE across these tasks. In contrast, \ourmethod achieves the best performance on both tasks, suggesting that its gains cannot be explained simply by increased diversity or conventional distributional regression. Full results and component ablations are reported in Appendix~\ref{sec:reward_ablation}.

\end{minipage}
\hfill
\begin{minipage}[t]{0.45\columnwidth}
\vspace{0pt}
\centering
\footnotesize
\setlength{\tabcolsep}{3.5pt}
\renewcommand{\arraystretch}{1.08}

\begin{tabular}{lcc}
\toprule
\textbf{Method} & \textbf{KBSS} $\uparrow$ & \textbf{LIPO} $\downarrow$ \\
\midrule
MSE & 0.60 & 0.822 \\
+ Variance & 0.59 & 0.807 \\
+ Entropy & 0.52 & 0.828 \\
\midrule
Gaussian NLL & 0.57 & 0.910 \\
Gaussian CRPS & 0.58 & 0.907 \\
Quantile / Pinball & 0.58 & 0.920 \\
\midrule
\textbf{\ourmethod} & \textbf{0.66} & \textbf{0.776} \\
\bottomrule
\end{tabular}

\captionof{table}{\textbf{Alternative Objectives.}}
\label{tab:alternative_objectives}

\end{minipage}

\section{Related Work}

\paragraph{LLM Regression.}
Recent work has explored LLMs as regressors through direct numeric prediction, in-context learning, embedding-based approaches, and regression-specific inference or fine-tuning~\citep{song2024omnipred,vacareanu2024words,nguyen2024predicting,tang2024understanding,yuchi2026llms,lukasik2024rail,lukasik2025better,chiang-etal-2025-tract,chen2025beyond}. Probabilistic approaches further elicit predictive distributions through quantile or numerical distribution prediction~\citep{vedula2025quantile,piskorz2026}, while recent benchmarks evaluate LLM regression across code, text, and model-performance prediction~\citep{akhauri2025regression,tchuindjo2025rir,park2025precog}. We extend this direction with on-policy RL using a distribution-aware reward.

\paragraph{RL and Distribution-Aware Reward Design.}
RL for LLM-based numeric prediction has primarily focused on forecasting, optimizing reasoning, accuracy, or calibration~\citep{luo2025time,koa2025reasoning,turtel2025outcome,chandak2025scaling}. Related work also considers uncertainty, calibration, and preserving output diversity during fine-tuning~\citep{harouni2026,damani2026beyond,petrenko2026entropy,he2025rewarding,xiao2025,anschel2025group}. In contrast, we study non-temporal scalar regression and optimize the empirical predictive distribution over numeric rollouts. Our leave-one-out credit assignment differs from RLOO~\citep{kool2019buy,ahmadian2024back}: rather than using leave-one-out rewards as an advantage baseline, we score each rollout’s contribution to distribution quality. Our approach also differs from ranking objectives~\citep{burges2005learning,joachims2002optimizing}, which optimize relative ordering rather than numeric predictive distributions.

\section{Conclusion}

We introduced \ourmethod, a distribution-aware RL objective that optimizes the predictive distribution formed by multiple LLM rollouts. 
Using CRPS with leave-one-out credit assignment, \ourmethod translates distribution-level quality into rollout-level rewards for on-policy RL. 
Across synthetic and real-world scientific regression tasks, \ourmethod improves point estimation and ranking while producing better-calibrated uncertainty estimates. 
Our analyses further show that distribution-aware training maintains meaningful variation across rollouts without explicitly maximizing dispersion. 
Together, these results highlight distribution-aware optimization as a promising direction for LLM regression with both accurate predictions and informative uncertainty.

\section*{Acknowledgements}

This research is supported in part by the NSF under grant numbers IIS-2052498, SMA-2418946, and NAIRR250217, in additon to a gift from Google. Any opinions, findings, and conclusions or recommendations expressed in this material are those of the author(s) and do not necessarily reflect the views of the National Science Foundation.

\bibliography{iclr2027_conference}
\bibliographystyle{iclr2027_conference}

\newpage


\appendix

\begin{figure}[t]
    \centering
    \includegraphics[width=1.0\columnwidth]{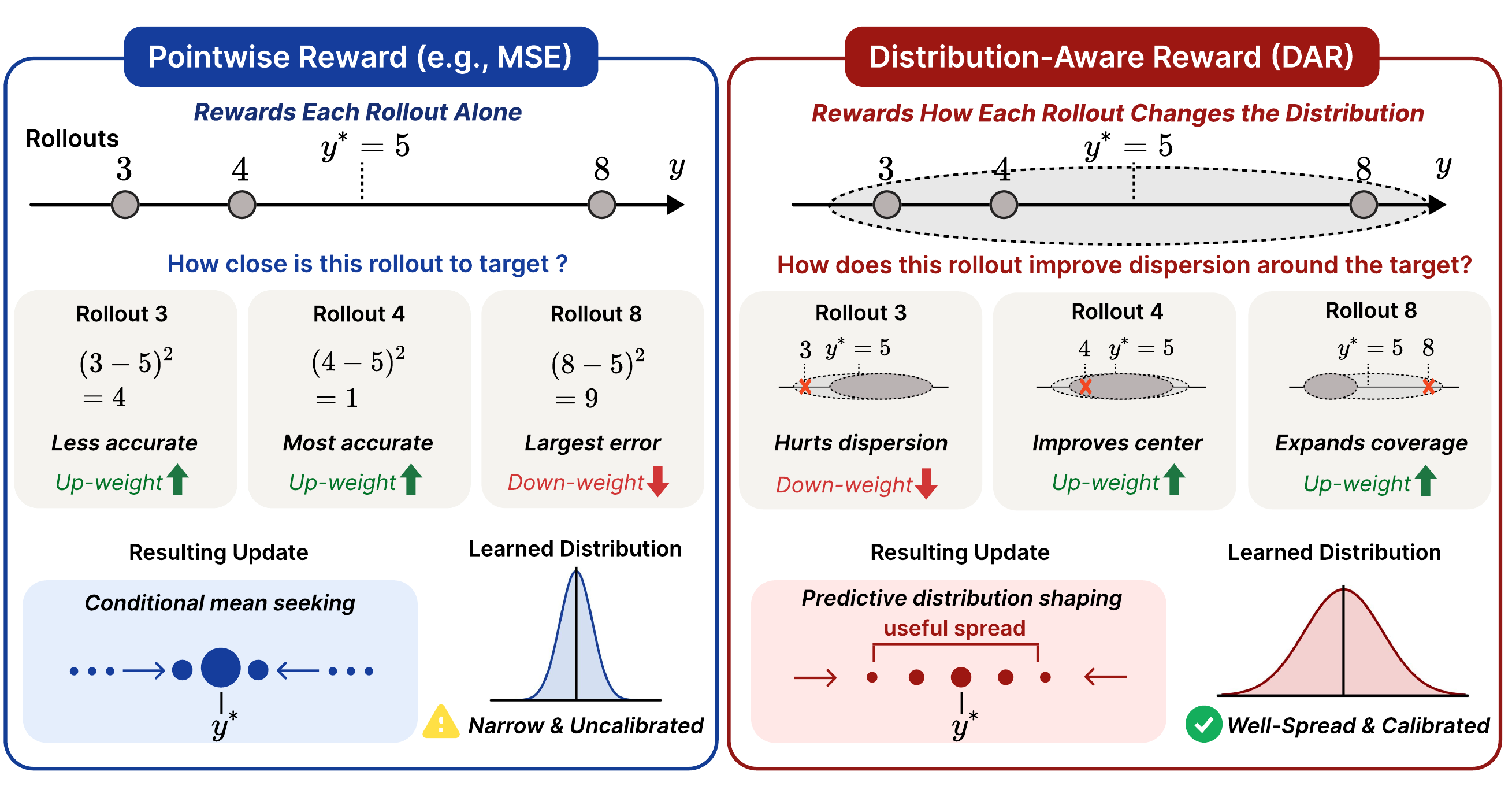}
    \caption{\textbf{Pointwise versus distribution-aware reward shaping.} Pointwise MSE reward scores each rollout independently, encouraging predictions to collapse toward the mean. \ourmethod instead scores each rollout by its leave-one-out contribution to the full predictive distribution, rewarding both accuracy and useful spread around the ground truth.}
    \label{fig:appendix_main_figure}
\end{figure}

\section{Proofs for Pointwise Squared Loss and CRPS}
\label{app:proof_mse_crps}

In this section, we formalize the difference between pointwise squared loss and CRPS at the population level. The key distinction is that pointwise squared loss evaluates a single prediction in isolation, whereas CRPS evaluates an entire predictive distribution.

\subsection{Theoretical Comparison of Pointwise Squared Loss and CRPS}

 The two objectives induce different optimal solutions: squared loss is mean-seeking, while CRPS is distribution-seeking~\citep{gneiting2011making, gneiting2007strictly, matheson1976scoring}.
These results are standard consequences of the Bayes optimality of squared loss for the conditional mean and the strict propriety of CRPS, but we restate them here to clarify the optimization bias induced by each objective in our multi-rollout regression setting.

\paragraph{Standard Result 1: Pointwise squared loss is mean-seeking.}
Fix an input $x$, and let $Y \mid X=x \sim P_x$ with finite second moment. Define the population pointwise squared-loss objective

\begin{equation}
\mathcal{L}_{\mathrm{MSE}}(a;x)
=
\mathbb{E}\big[(a-Y)^2 \mid X=x\big],
\qquad a \in \mathbb{R}.
\end{equation}

Then $\mathcal{L}_{\mathrm{MSE}}(a;x)$ is uniquely minimized at

\begin{equation}
a^\star(x)=\mu_x:=\mathbb{E}[Y\mid X=x].
\end{equation}

Equivalently, among predictive distributions over scalar predictions, the minimizer of expected pointwise squared loss is the degenerate predictor $\delta_{\mu_x}$ concentrated at the conditional mean~\citep{gneiting2011making}.

\paragraph{Proof.}
Let $\mu_x = \mathbb{E}[Y \mid X=x]$. Expanding the square around $\mu_x$ gives

\begin{align}
\mathcal{L}_{\mathrm{MSE}}(a;x)
&=
\mathbb{E}\big[(a-Y)^2 \mid X=x\big] \\
&=
\mathbb{E}\big[(a-\mu_x+\mu_x-Y)^2 \mid X=x\big] \\
&=
(a-\mu_x)^2
+2(a-\mu_x)\mathbb{E}[\mu_x-Y\mid X=x]
+\mathbb{E}\big[(\mu_x-Y)^2\mid X=x\big].
\end{align}

Since $\mathbb{E}[\mu_x-Y\mid X=x]=0$, this simplifies to

\begin{equation}
\mathcal{L}_{\mathrm{MSE}}(a;x)
=
(a-\mu_x)^2+\mathrm{Var}(Y\mid X=x).
\end{equation}

The variance term does not depend on $a$, so the unique minimizer is $a=\mu_x$.

Now consider a predictive distribution $F$ over scalar predictions $\hat{Y}$. If the model is evaluated by pointwise squared loss on a single sampled prediction $\hat{Y}\sim F$, then the population objective becomes

\begin{align}
\mathbb{E}_{\hat{Y}\sim F}\,\mathbb{E}\big[(\hat{Y}-Y)^2\mid X=x\big]
&=
\mathbb{E}_{\hat{Y}\sim F}\left[(\hat{Y}-\mu_x)^2+\mathrm{Var}(Y\mid X=x)\right] \\
&=
\mathbb{E}_{\hat{Y}\sim F}[(\hat{Y}-\mu_x)^2]+\mathrm{Var}(Y\mid X=x).
\end{align}

This is minimized by placing all mass at $\mu_x$, namely $F=\delta_{\mu_x}$. Therefore, under pointwise squared loss, any predictive spread away from the conditional mean strictly increases the objective. This establishes that pointwise squared loss is inherently mean-seeking and favors collapse to a point mass, which is empirically verified in our analysis.

\paragraph{Standard Result 2: CRPS is distribution-seeking.}
Fix an input $x$, and let $Y \mid X=x \sim P_x$. For any predictive distribution $F$ on $\mathbb{R}$ with finite first moment, define

\begin{equation}
\mathcal{L}_{\mathrm{CRPS}}(F;x)
=
\mathbb{E}_{Y\sim P_x}\big[\mathrm{CRPS}(F,Y)\big].
\end{equation}

Then $\mathcal{L}_{\mathrm{CRPS}}(F;x)$ is minimized at $F=P_x$. In particular, if $P_x$ is non-degenerate, then for any point mass $\delta_a$,

\begin{equation}
\mathcal{L}_{\mathrm{CRPS}}(P_x;x)
<
\mathcal{L}_{\mathrm{CRPS}}(\delta_a;x).
\end{equation}

Hence, when $\mathrm{Var}(Y\mid X=x)>0$, collapse to any single point, including the conditional mean, is suboptimal under CRPS~\citep{matheson1976scoring, gneiting2007strictly}.

\paragraph{Proof.}
CRPS is a strictly proper scoring rule for univariate predictive distributions~\citep{matheson1976scoring, gneiting2007strictly}. Therefore, for any true conditional distribution $P_x$ and any predictive distribution $F$ with finite first moment,

\begin{equation}
\mathbb{E}_{Y\sim P_x}\big[\mathrm{CRPS}(P_x,Y)\big]
\le
\mathbb{E}_{Y\sim P_x}\big[\mathrm{CRPS}(F,Y)\big],
\end{equation}

with equality if and only if $F=P_x$.

This immediately proves the first claim: the population CRPS objective is minimized by the true conditional predictive distribution.

For the second claim, suppose $P_x$ is non-degenerate. Then $P_x \neq \delta_a$ for every $a\in\mathbb{R}$. By strict propriety of CRPS, any predictive distribution different from $P_x$ must incur strictly larger expected CRPS. In particular,

\begin{equation}
\mathcal{L}_{\mathrm{CRPS}}(P_x;x)
<
\mathcal{L}_{\mathrm{CRPS}}(\delta_a;x)
\qquad
\text{for all } a\in\mathbb{R}.
\end{equation}

Applying this to $a=\mu_x=\mathbb{E}[Y\mid X=x]$ shows that the mean-collapsed predictor $\delta_{\mu_x}$ is strictly suboptimal whenever the conditional target distribution has nonzero spread.

Thus, unlike pointwise squared loss, CRPS does not generally favor collapse to the conditional mean. Rather, its population optimum matches the true conditional distribution: it remains non-degenerate when the conditional target distribution is non-degenerate, and reduces to a point mass at the ground-truth target when the conditional target is deterministic. 
As discussed in Section~\ref{app:dar_deterministic}, the benefit of DAR in the deterministic setting therefore arises not from changing the population optimum, but from providing a distribution-aware learning signal over finite stochastic rollouts.

\subsection{Implication}

The two standard results above formalize the contrast between pointwise and distributional regression objectives at the population level.
Pointwise squared loss treats each prediction independently and is optimized by the conditional mean, so it provides no incentive to represent predictive uncertainty.
In contrast, CRPS is a strictly proper scoring rule for predictive distributions and is optimized by the true conditional predictive law when that law is observed through repeated samples.

This population-level result motivates CRPS as a distribution-aware scoring rule.
In our finite-sample LLM setting, however, DAR does not directly observe the full conditional distribution for each input.
Instead, it uses the empirical negated CRPS against the available scalar target as a reward over multiple sampled rollouts.
Thus, DAR should be viewed as a practical distribution-aware credit-assignment objective that encourages well-centered and appropriately dispersed rollout sets, rather than as a guarantee of recovering the true conditional distribution.

\section{Distribution-Aware Rewards for Deterministic Regression}
\label{app:dar_deterministic}

Even when each input has a single deterministic target, an LLM naturally induces a predictive distribution through multiple numerical rollouts due to finite data, model approximation, and stochastic generation. This raises a natural question: why should optimizing this distribution help when the underlying regression target itself is deterministic?

A conventional pointwise reward such as MSE evaluates each rollout independently, encouraging sampled predictions to concentrate around a single point. In contrast, \ourmethod evaluates each rollout by how it changes the quality of the empirical predictive distribution formed by all sampled rollouts. As illustrated in Figure~\ref{fig:main_figure}, a rollout can therefore receive useful credit not only for being individually close to the target, but also for improving the distribution as a whole---for example, by shifting its center toward the target or placing predictions on both sides of the target. Conversely, rollouts that introduce unhelpful dispersion are penalized. Thus, \ourmethod encourages a predictive distribution that is accurate and appropriately dispersed, rather than diversity for its own sake.

This distributional supervision can also improve the resulting point estimate. Our final prediction is obtained by averaging multiple sampled outputs. When the sampled predictions fall on both sides of the target, positive and negative errors can partially cancel under averaging, reducing the error of the aggregate prediction. Consistent with this mechanism, \ourmethod increases the bracketing rate---the fraction of examples for which the rollout set contains predictions on both sides of the ground-truth target---from 37.7\% to 53.6\% (Figure~\ref{fig:mae_bracket_comparison}). This provides empirical evidence for how distribution-level supervision can improve point-prediction accuracy even when the target is deterministic.

Importantly, \ourmethod does not encourage predictive spread for its own sake. CRPS jointly accounts for accuracy and dispersion, so additional spread is beneficial only when it improves the predictive distribution relative to the observed target. As predictions approach the ground truth, unnecessary dispersion is increasingly penalized. The role of \ourmethod is therefore to provide a richer optimization signal during learning, rather than to maintain persistent uncertainty at convergence.

This intuition does not require stochastic or repeated labels. CRPS is a strictly proper scoring rule that can be evaluated against a single observed target~\citep{gneiting2007strictly}. For a predictive distribution $F$ and target $y$,

\begin{equation}
    \mathrm{CRPS}(F,y)
    =
    \mathbb{E}|X-y|
    -
    \frac{1}{2}\mathbb{E}|X-X'|,
    \qquad X,X'\sim F.
\end{equation}

By the triangle inequality,

\begin{equation}
    \frac{1}{2}\mathbb{E}|X-X'|
    \le
    \mathbb{E}|X-y|,
\end{equation}

and therefore

\begin{equation}
    \mathrm{CRPS}(F,y)\ge0.
\end{equation}

Equality is attained when $F=\delta_y$, such that all predictive mass is concentrated at the correct target. Hence, for deterministic regression, the global optimum under CRPS remains a point mass at the ground-truth target. \ourmethod therefore does not alter the desired solution; rather, it changes the optimization signal and trajectory used to reach it.

\section{Estimator Properties of Leave-One-Out-CRPS}
\label{sec:loo_crps_properties}

Our method uses LOO-CRPS to assign a distribution-aware reward to each rollout based on its contribution to the empirical predictive distribution. Because this reward is computed from a finite set of sampled rollouts, two theoretical questions naturally arise: how the leave-one-out (LOO) credit assignment affects the resulting policy-gradient estimator, and whether empirical CRPS introduces finite-sample bias. We analyze both properties below.

\paragraph{Leave-One-Out credit assignment.}
LOO credit assignment does not introduce additional bias into the policy-gradient estimator. For a group of $K$ rollouts $\{x_1,\ldots,x_K\}$, the LOO baseline for rollout $x_k$ is computed exclusively from the remaining $K-1$ rollouts. Conditional on the input, this baseline is therefore independent of the stochastic action corresponding to rollout $x_k$. As in standard multi-sample LOO estimators~\citep{kool2019buy}, subtracting such a baseline does not change the expected REINFORCE gradient:

\begin{equation}
    \mathbb{E}_{x_k \sim \pi_\theta}
    \left[
        b_{-k}\nabla_\theta \log \pi_\theta(x_k)
    \right]
    =
    b_{-k}
    \mathbb{E}_{x_k \sim \pi_\theta}
    \left[
        \nabla_\theta \log \pi_\theta(x_k)
    \right]
    = 0,
\end{equation}

where $b_{-k}$ denotes a quantity computed only from the other $K-1$ rollouts. LOO credit assignment can therefore provide rollout-specific credit while preserving the expected policy-gradient estimator.

\paragraph{Finite-sample properties of empirical CRPS.}
The empirical CRPS computed from a finite rollout set is a biased but consistent estimator of the population CRPS. For $K$ i.i.d.\ samples $X_1,\ldots,X_K \sim F$, the empirical CRPS can be written as

\begin{equation}
    \widehat{\mathrm{CRPS}}_K
    =
    \frac{1}{K}\sum_{k=1}^{K}|X_k-y|
    -
    \frac{1}{2K^2}
    \sum_{k=1}^{K}\sum_{j=1}^{K}|X_k-X_j|.
\end{equation}

The first term is an unbiased estimator of $\mathbb{E}|X-y|$. For the pairwise dispersion term, the $K$ diagonal terms are zero, leaving $K(K-1)$ nonzero pairs in expectation. Consequently,

\begin{equation}
    \mathbb{E}
    \left[
        \frac{1}{2K^2}
        \sum_{k=1}^{K}\sum_{j=1}^{K}|X_k-X_j|
    \right]
    =
    \frac{K-1}{K}
    \frac{1}{2}\mathbb{E}|X-X'|,
\end{equation}

where $X,X'\sim F$ are independent. The empirical estimator therefore underestimates the dispersion term by a factor of $(K-1)/K$, resulting in an $O(1/K)$ finite-sample bias.

As $K\rightarrow\infty$, this factor converges to one, and the empirical CRPS converges to the population CRPS. Correspondingly, the LOO contribution increasingly approximates the marginal effect of each rollout on the population objective. Importantly, strict propriety is a property of the population CRPS and does not imply that each finite-$K$ LOO reward is itself a proper scoring rule. At finite $K$, the empirical objective is slightly conservative with respect to predictive dispersion; this bias vanishes as the number of rollouts increases.

Together, these properties distinguish two aspects of the estimator: LOO credit assignment does not introduce additional policy-gradient bias, while empirical CRPS has a finite-sample $O(1/K)$ bias that vanishes asymptotically.

\section{Experimental Setup}
\label{sec:appendix_experimental_setup}

In this section, we provide additional details on the experimental setup for the synthetic benchmark and the two real-world regression datasets, as well as the implementation details and baseline methods.

\subsection{Tasks}
\label{subsec:appendix_tasks}

\paragraph{Synthetic Distributional Regression.}
We construct a synthetic 1D distributional regression benchmark following Appendix E.2 of \citet{fishkov2025uncertainty}. For each input \(x \in \mathbb{R}\), the ground-truth conditional distribution is a heteroscedastic two-component mixture:

\begin{equation}
p^{*}(y \mid x) = \pi(x) P_1(y \mid x) + \bigl(1-\pi(x)\bigr) P_2(y \mid x).
\end{equation}

The input-dependent mixing weight is defined as

\begin{equation}
\pi(x)=\frac{1}{1+\exp(1.2x)},
\end{equation}

and the two component means are given by

\begin{equation}
\mu_1(x)=\frac{x}{3}+1.2\sin(0.8x),
\end{equation}

\begin{equation}
\mu_2(x)=\frac{x}{3}-1.2\cos(0.8x).
\end{equation}

The input-dependent noise scale is

\begin{equation}
\sigma(x)=0.12+0.28\left(0.5+0.5\sin(0.7x)\right)^2.
\end{equation}

Each component is Gaussian with shared heteroscedastic variance:

\begin{equation}
P_1(y \mid x)=\mathcal{N}\!\left(y;\mu_1(x),\sigma^2(x)\right),
\end{equation}

\begin{equation}
P_2(y \mid x)=\mathcal{N}\!\left(y;\mu_2(x),\sigma^2(x)\right).
\end{equation}

Equivalently, one may sample from the distribution via

\begin{equation}
k \sim \mathrm{Bernoulli}(\pi(x)),
\end{equation}

\begin{equation}
\epsilon \sim \mathcal{N}(0,1),
\end{equation}

\begin{equation}
y=\mu_k(x)+\epsilon\,\sigma(x).
\end{equation}

\begin{figure}[t]
    \centering
    \includegraphics[width=0.6\columnwidth]{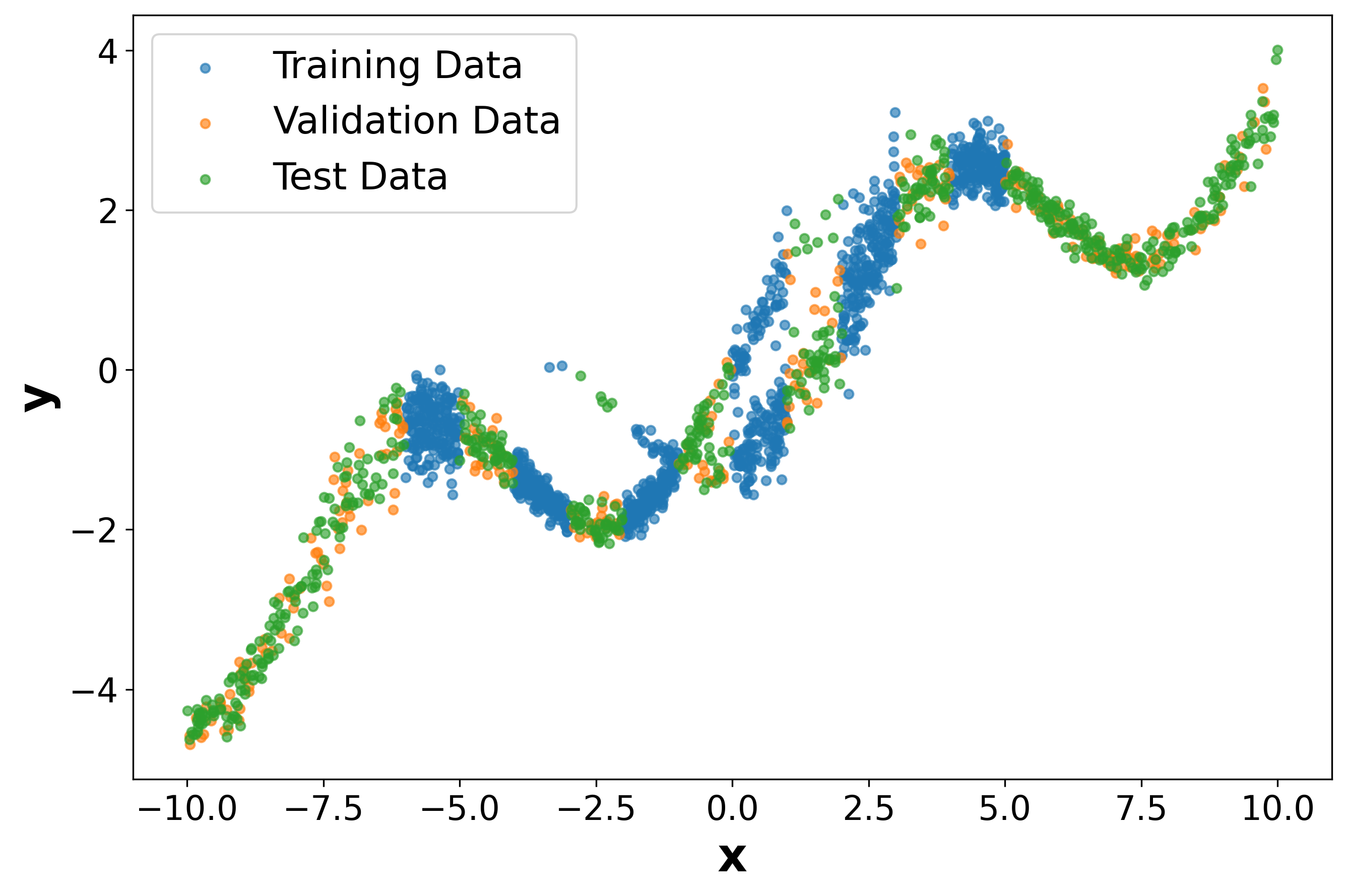}
        \caption{Illustration of data samples}
    \label{fig:synthetic_data_samples}
\end{figure}

Following the same setup, we draw \(n=1200\) training pairs \((x,y)\) from \(p^{*}(y \mid x)\). In our implementation, training inputs are sampled from the in-domain interval \(x\in[-6,6]\), while the test set covers both the training region and additional disjoint regions outside \([-6,6]\) in order to evaluate both interpolation and extrapolation behavior. Figure~\ref{fig:synthetic_data_samples} shows the one-dimensional distribution of the data samples.

\paragraph{Coding Performance Prediction.}
For code performance prediction, we use the \texttt{akhauriyash/Code-Regression} benchmark and restrict attention to the APPS and KBSS portions.\footnote{The Hugging Face dataset card lists the dataset license as MIT.} Following \citet{akhauri2025regression}, we construct random train/validation/test splits over these subsets. We adopt this protocol because the original authors do not provide an official fixed split for these data, and thus the random-split setup is consistent with their experimental practice. The task is to predict a continuous execution-related target from code alone, and our evaluation follows the standard regression setting used in prior work. The released dataset card specifies an MIT license for the packaged benchmark.

\paragraph{MoleculeNet Property Prediction.}
For molecular property prediction, we use the MoleculeNet benchmark \citep{wu2018moleculenet} and focus on the three commonly used physicochemical regression subsets: ESOL, FreeSolv, and Lipophilicity (Lipo). MoleculeNet explicitly includes these regression tasks and recommends scaffold-based evaluation for such chemistry benchmarks, which is also the protocol commonly followed in recent work~\citep{zhou2023unimol, yu2024llasmol}.  
In line with prior molecular representation work, including Uni-Mol and subsequent studies, we use the official scaffold split for ESOL, FreeSolv, and Lipo.  

Regarding licensing, MoleculeNet is a benchmark that aggregates public molecular datasets, while the DeepChem software commonly used to access these benchmark tasks is released under the MIT License.  
Since ESOL, FreeSolv, and Lipophilicity originate from public source datasets collected into the MoleculeNet benchmark, usage should also respect the terms of the original underlying sources in addition to the benchmark software license. This benchmark structure is documented in MoleculeNet and related dataset documentation.  

\subsection{Baselines}

For zero-shot evaluation of domain-specific language models, we follow the default generation protocol described in each original paper~\citep{fang2023mol, zhang2024chemllm, yu2024llasmol}.
For ICL~\citep{vacareanu2024words}, we use 50 in-context examples for MoleculeNet property prediction and 10 for coding regression due to context length constraints. Following the original setup, these examples are randomly sampled from the training set.

For the value-head baseline, we follow the embedding-based regression setup of \citet{tang2024understanding}. Specifically, we keep the language model frozen and train only a lightweight prediction head on top of fixed input representations. For each example, we extract a fixed-dimensional embedding from the \texttt{prompt} field using a pretrained Qwen instruction model, and use this embedding as input to a two-hidden-layer MLP regressor with ReLU activations. The regression head is trained with mean squared error (MSE) on normalized target values, where the normalization statistics are computed only from the training split. At evaluation time, predictions are transformed back to the original target scale before computing all regression metrics. We optimize the regressor with AdamW, select the best checkpoint based on validation performance, and use early stopping with a maximum of 300 training steps. Overall, this setup closely follows the protocol of \citet{tang2024understanding}.

For both SFT and RL training, we use the default training configuration provided in the \texttt{verl} repository. We keep the batch size and the number of training steps the same across methods for fair comparison.

\subsection{Computing Resources}
\label{subsec:computing_resources}

We use a single node with 4 H100 GPUs for all SFT and RL training runs. Depending on the model scale and the maximum number of training steps, each run takes approximately 1–2 days. Our method does not introduce additional computational overhead relative to the pointwise GRPO baseline.

\subsection{Implementation Details}
\label{sec:appendix_implementation}

\begin{table*}[t]
\centering
\small
\begin{tabular}{ll}
\toprule
\textbf{Category} & \textbf{Value} \\
\midrule
Framework & \texttt{verl} \\
Training method & GRPO \\
Train batch size & 256 \\
Learning rate & $1\times10^{-6}$ \\
PPO mini-batch size & 16 \\
PPO micro-batch size / GPU & 4 \\
Entropy coefficient & 0 \\
Use KL loss & True \\
KL loss coefficient & 0.001 \\
KL loss type & \texttt{low\_var\_kl} \\
Use KL in reward & False \\
Rollout engine & vLLM \\
Rollouts per prompt ($n$) & 12 \\
Sampling temperature & 1.0 \\
Top-$p$ & 1.0 \\
Rollout GPU memory utilization & 0.8 \\
Rollout log-prob micro-batch size / GPU & 8 \\
Reference log-prob micro-batch size / GPU & 8 \\
Gradient checkpointing & True \\
Parameter offload & False \\
Optimizer offload & False \\
Critic warmup & 0 \\
GPUs & 4 \\
Nodes & 1 \\
\bottomrule
\end{tabular}
\caption{\textbf{Main RL hyperparameters for DAR training in \texttt{verl}.} Our method follows the standard GRPO pipeline and differs mainly in the reward computation.}
\label{tab:impl_hparams}
\end{table*}

Table~\ref{tab:impl_hparams} shows the detailed setup of the training.
We implement both supervised fine-tuning (SFT) and reinforcement learning (RL) using the \texttt{verl}\footnote{https://github.com/verl-project/verl} framework. For RL, we build on the standard GRPO training pipeline and replace only the reward computation with our distribution-aware reward. Concretely, for each prompt, the policy generates multiple rollouts, and our reward manager computes rollout-level rewards based on each sample's contribution to the predictive distribution induced by the full rollout set. This design requires no additional value head, no auxiliary uncertainty head, and no extra learned scorer beyond the standard policy/reference setup.

Our method introduces only minimal computational overhead relative to standard GRPO. The dominant training cost remains rollout generation and policy optimization, which are unchanged. The additional cost comes only from reward-side aggregation over the sampled rollouts for each prompt. In practice, this is lightweight compared to model forward passes, since it operates on scalar predictions after decoding rather than on token-level hidden states. Thus, the training pipeline, hardware footprint, and optimization procedure remain nearly identical to standard GRPO. We trained each model on each dataset for sufficiently long training, up to a maximum of 300 steps, selected the best checkpoint based on validation performance, and then reported the final results on the test set.

\subsection{Computational Overhead}
\label{sec:computational_overhead}

\ourmethod introduces only negligible computational overhead relative to standard negative-MSE RL under matched GRPO-style training settings. Importantly, multiple rollouts are not an additional cost introduced by \ourmethod: both the negative-MSE baseline and \ourmethod operate on the same $K=12$ rollouts sampled per prompt. The two methods therefore share the same rollout generation, log-probability evaluation, and policy optimization procedures, differing only in how rewards are computed from the sampled predictions.

\begin{table}[h]
\small
\centering
\begin{tabular}{lccc}
\toprule
\textbf{Metric} & \textbf{Negative-MSE} & \textbf{\ourmethod} & \textbf{Overhead} \\
\midrule
Reward computation (ms/step) & $\sim$15 & 83 & +56--73 ms \\
Share of a 207.9 s training step & $<$0.01\% & 0.04\% & +0.03--0.04\% \\
Reward computation at $K=48$ (ms) & -- & 302 & $<$0.2\% of step \\
\bottomrule
\end{tabular}
\caption{\textbf{Computational overhead analysis.} Computational overhead of \ourmethod relative to negative-MSE RL using Qwen3-4B with $K=12$ rollouts under matched settings.}
\label{table:runtime_overhead}
\end{table}

As shown in Table~\ref{table:runtime_overhead}, the entire reward-processing stage for \ourmethod accounts for only 0.04\% of a 207.9\,s training step. Relative to negative-MSE RL, \ourmethod adds only 56--73\,ms per step, corresponding to approximately 0.03--0.04\% additional wall-clock time. We also observe no measurable increase in peak GPU memory.

The additional computation arises from the pairwise CRPS evaluation over the $K$ predictions generated for each prompt. This computation has theoretical complexity $O(K^2)$, but it is restricted to reward evaluation and is negligible compared with rollout generation and policy optimization. Increasing the rollout count from $K=12$ to $K=48$ increases the measured reward-computation time from 83\,ms to 302\,ms, which still accounts for less than 0.2\% of a typical training step. Because the measured runtime also includes fixed reward-processing overhead, the observed scaling need not follow the asymptotic pairwise complexity exactly.

Overall, \ourmethod changes the reward computation applied to rollouts that are already required by GRPO-style training, while leaving the dominant computational components unchanged. Its distribution-aware objective therefore incurs negligible practical runtime and memory overhead.

\section{Distribution Characteristics}
\label{app:crps}

\begin{table*}[t]
\centering
\footnotesize
\setlength{\tabcolsep}{6pt}
\renewcommand{\arraystretch}{1.08}
\begin{tabular}{llrrrr}
\toprule
\textbf{Dataset} & \textbf{Method} &
\textbf{CRPS}$\downarrow$ &
\textbf{Predictive SD} &
\textbf{IQR} &
\textbf{Skewness} \\
\midrule
KBSS
& SFT & 0.307 & 0.014 & 0.016 & 1.164 \\
& MSE & 0.411 & 0.001 & 0.001 & -0.447 \\
\rowcolor[HTML]{FEE2E2} & \ourmethod & 0.339 & 0.005 & 0.007 & 0.279 \\
\midrule
APPS
& SFT & 0.071 & 2070.000 & 1067.250 & 2.082 \\
& MSE & 0.072 & 0.002 & 0.002 & 0.004 \\
\rowcolor[HTML]{FEE2E2} & \ourmethod & 0.070 & 303.279 & 296.402 & 0.399 \\
\midrule
ESOL
& SFT & 0.307 & 0.171 & 0.171 & 0.171 \\
& MSE & 0.296 & 0.157 & 0.171 & -0.300 \\
\rowcolor[HTML]{FEE2E2} & \ourmethod & 0.279 & 0.255 & 0.305 & -0.121 \\
\midrule
FreeSolv
& SFT & 0.460 & 0.317 & 0.371 & -0.205 \\
& MSE & 0.478 & 0.179 & 0.166 & 0.878 \\
\rowcolor[HTML]{FEE2E2} & \ourmethod & 0.420 & 0.311 & 0.350 & 0.051 \\
\midrule
LIPO
& SFT & 0.505 & 0.219 & 0.249 & -0.246 \\
& MSE & 0.561 & 0.056 & 0.049 & -0.160 \\
\rowcolor[HTML]{FEE2E2} & \ourmethod & 0.498 & 0.128 & 0.162 & -0.045 \\
\bottomrule
\end{tabular}
\caption{\textbf{Predictive distribution quality across datasets.}
CRPS is the primary metric, where lower values are better. Predictive SD, IQR, and skewness characterize the spread and shape of the raw predictive samples generated by each checkpoint. CRPS values are normalized within each dataset, while SD and IQR are reported in the original target units. All values are rounded to three decimal places.}
\label{tab:crps_distribution}
\end{table*}

Table~\ref{tab:crps_distribution} reports CRPS together with complementary statistics describing the predictive distributions. \ourmethod achieves the best CRPS on four of the five datasets and consistently outperforms the MSE baseline. The only exception is KBSS, where SFT obtains a lower CRPS.
Moreover, \ourmethod generally produces less asymmetric predictive distributions than SFT, with skewness values closer to zero across datasets.

On KBSS, \ourmethod achieves better point prediction accuracy than SFT but produces a substantially narrower predictive distribution. This concentration makes \ourmethod more sensitive to a small subset of difficult tail examples, which increases its average CRPS despite better performance on most examples. When the worst tail examples are excluded (2\%), the ranking reverses and \ourmethod slightly outperforms SFT. Thus, the KBSS result is best understood as a dataset-specific effect related to the concentration of the predictive distribution, rather than a systematic disadvantage of \ourmethod.

\section{Reward and Credit Assignment Ablations}
\label{sec:reward_ablation}

\begin{table}[t]
\centering
\footnotesize
\setlength{\tabcolsep}{3.5pt}
\renewcommand{\arraystretch}{1.05}
\begin{tabular}{lccccc}
\toprule
\textbf{Method}
& \textbf{KBSS} $\uparrow$
& \textbf{APPS} $\uparrow$
& \textbf{FreeSolv} $\downarrow$
& \textbf{LIPO} $\downarrow$
& \textbf{ESOL} $\downarrow$ \\
\midrule
MSE
& 0.60 & 0.92 & 2.561 & 0.822 & 0.878 \\
+ Variance
& 0.59 & 0.93 & 2.388 & 0.807 & 0.881 \\
+ Entropy
& 0.52 & 0.93 & 2.424 & 0.828 & 0.885 \\
\midrule
Gaussian NLL
& 0.57 & 0.78 & 3.389 & 0.910 & 0.877 \\
Gaussian CRPS
& 0.58 & 0.67 & 2.866 & 0.907 & 0.883 \\
Quantile / Pinball
& 0.58 & 0.62 & 2.972 & 0.920 & 0.890 \\
\midrule
CRPS + Group-Mean
& 0.62 & 0.94 & 2.374 & 0.781 & 0.875 \\
NLL + LOO
& 0.63 & 0.95 & 2.368 & 0.797 & 0.880 \\
\midrule
\rowcolor[HTML]{FEE2E2} \textbf{\ourmethod}
& \textbf{0.66}
& \textbf{0.96}
& \textbf{2.360}
& \textbf{0.776}
& \textbf{0.871} \\
\bottomrule
\end{tabular}
\captionof{table}{
\textbf{Alternative objectives and component ablations.} We compare \ourmethod with diversity-regularized MSE objectives, standard distributional regression objectives, and ablations of its CRPS objective and LOO credit assignment. Spearman correlation is reported for KBSS and APPS ($\uparrow$), and RMSE for molecular property prediction tasks ($\downarrow$). Best results are shown in bold.}
\label{tab:alternative_objectives_full}
\end{table}

We compare \ourmethod against several simpler alternatives, including negative-MSE objectives with variance or entropy bonuses, standard distributional regression objectives, and component ablations that replace either the CRPS objective or LOO credit assignment. As shown in Table~\ref{tab:alternative_objectives_full}, \ourmethod achieves the best performance on all five datasets, outperforming both diversity-regularized and conventional distributional baselines as well as either component ablation. These results indicate that the gains do not arise from predictive diversity, distributional training, or either component alone, but from combining empirical distribution-level evaluation with rollout-specific LOO credit assignment.

\section{Sensitivity to Hyperparameters}
\label{sec:hyperparameter_sensitivity}

We examine the sensitivity of \ourmethod to two sampling hyperparameters: the number of rollouts $K$ and the sampling temperature $T$. We evaluate rollout counts $K\in\{4,8,12,24\}$ and sampling temperatures $T\in\{0.7,1.0,1.3\}$ on KBSS and the synthetic regression task.

\begin{table}[t]
\centering
\small
\begin{tabular}{lcccc}
\toprule
\textbf{Dataset} & $\mathbf{K=4}$ & $\mathbf{K=8}$ & $\mathbf{K=12}$ & $\mathbf{K=24}$ \\
\midrule
KBSS      & 0.59 & 0.65 & 0.66 & 0.66 \\
Synthetic & 1.78 & 1.40 & 1.49 & 1.42 \\
\bottomrule
\end{tabular}
\caption{\textbf{Sensitivity to the number of rollouts $K$.} We report Spearman correlation ($\uparrow$) on KBSS and RMSE ($\downarrow$) on the synthetic task.}
\label{tab:rollout_sensitivity}
\end{table}

\begin{table}[t]
\centering
\small
\begin{tabular}{lccc}
\toprule
\textbf{Dataset} & $\mathbf{T=0.7}$ & $\mathbf{T=1.0}$ & $\mathbf{T=1.3}$ \\
\midrule
KBSS      & 0.62 & 0.66 & 0.65 \\
Synthetic & 1.55 & 1.49 & 1.48 \\
\bottomrule
\end{tabular}
\caption{\textbf{Sensitivity to sampling temperature $T$.} We report Spearman correlation ($\uparrow$) on KBSS and RMSE ($\downarrow$) on the synthetic task.}
\label{tab:temperature_sensitivity}
\end{table}

Tables~\ref{tab:rollout_sensitivity} and~\ref{tab:temperature_sensitivity} show that \ourmethod is relatively robust to both hyperparameters. Performance generally improves as the number of rollouts increases and stabilizes beyond $K=8$, with little additional benefit from increasing $K$ to 24. Similarly, varying the sampling temperature from $0.7$ to $1.3$ results in only modest performance differences across both tasks. These results indicate that the performance of \ourmethod does not depend strongly on a particular choice of rollout count or sampling temperature.

\section{Reasoning Mode vs. Non-Reasoning Mode}
\label{sec:analysis-reasoningvsnon_reasoning}

\begin{table*}[t]
\centering
\small
\setlength{\tabcolsep}{8pt}
\begin{tabular}{cccccccc}
\toprule
\textbf{Model} & \textbf{Reasoning} &
\multicolumn{3}{c}{\textbf{Zero-shot}} &
\multicolumn{3}{c}{\textbf{DAR (ESOL)}} \\
\cmidrule(lr){3-5}\cmidrule(lr){6-8}
& & \textbf{\# Tokens} & $\rho\,(\uparrow)$ & RMSE\,($\downarrow$) & \textbf{\# Tokens} & $\rho\,(\uparrow)$ & RMSE\,($\downarrow$) \\
\midrule
Qwen3-4B & No  & 1072 & 0.67 & 1.62 & 7 & 0.90 & 0.73 \\
Qwen3-4B & Yes & 2737 & 0.66 & 1.68 & 2584 & 0.87 & 1.01 \\
\bottomrule
\end{tabular}
\caption{\textbf{ESOL regression results with and without explicit reasoning.} We report the average number of output tokens, Spearman correlation ($\rho$), and RMSE for zero-shot and DAR.}
\label{tab:esol_reasoning_train_only}
\end{table*}

On-policy RL is naturally compatible with reasoning-oriented models because it can reinforce longer intermediate derivations. We nevertheless disable explicit reasoning by default in our main experiments for efficiency, and instead study its effect through separate ablations in both the zero-shot and RL fine-tuning settings. In these ablations, we explicitly instruct the model in the system prompt to reason before producing a prediction, and set the maximum output length to 8192 during training.

Table~\ref{tab:esol_reasoning_train_only} presents a controlled comparison on ESOL using Qwen3-4B-Instruct, evaluating both \texttt{Reasoning=Yes} and \texttt{Reasoning=No} under two settings: (i) zero-shot inference and (ii) DAR fine-tuning on ESOL. The results show that explicitly prompting the model to generate intermediate reasoning~\citep{wei2022chain} substantially increases the number of output tokens, but does not improve regression performance in either setting. In the fine-tuning case, reasoning is in fact associated with worse performance. This trend is consistent with recent findings in GRPO training for verifiable tasks, which suggest that RL-based reasoning training can reduce output diversity~\citep{he2025rewarding, petrenko2026entropy}. A similar effect may be present here, as the reasoning-enabled model exhibits lower variance than the non-reasoning counterpart (total variance 3.5 vs.\ 4.2). Note that this finding is specific to our single-task fine-tuning setup on ESOL with small scale model, where training and evaluation are performed on the same task. It does not rule out the possibility that explicit reasoning could help in broader settings such as multi-task post-training or transfer.

\paragraph{Reward scale.}
\ourmethod also does not require task-specific reward scaling despite differences in target ranges across datasets. Under GRPO, advantages are normalized by the within-group standard deviation, making the policy-gradient signal invariant to affine rescaling of the reward. Consequently, differences in the absolute magnitude of CRPS across target ranges do not alter the effective reward scale used for policy optimization. Consistent with this property, we use the same reward formulation and KL coefficient across all five datasets without per-task tuning.

\section{Limitations}
\label{sec:limitations}

\paragraph{Comparison to traditional regression methods.}
While our results suggest that LLM-based regression can be useful for scientific prediction tasks, this approach also has important limitations relative to traditional nonlinear regression methods. Classical models such as Gaussian processes, kernel methods, gradient-boosted trees, and neural regression models are often more computationally efficient, easier to train, and better suited when the input is already available as structured numerical or tabular features. They also provide clearer inductive biases and, in some cases, more established uncertainty estimates. In contrast, LLM-based regression requires substantially more compute, depends on careful prompting and decoding choices, and may be less sample-efficient when the task does not benefit from language-based representations.

\paragraph{Flexibility and sensitivity.}
The main advantage of LLM-based regression is its flexibility: the same model can consume heterogeneous inputs such as natural-language descriptions, code, SMILES strings, metadata, or task specifications without task-specific feature engineering. This makes it attractive for settings where the input is naturally textual or where constructing specialized features is costly. However, this flexibility also introduces additional sources of variability. Predictions can be sensitive to prompt format, sampling temperature, output parsing, and the choice of backbone model. Moreover, although our distribution-aware objective improves rollout diversity and calibration, it does not eliminate all failure modes of RL training, such as unstable optimization, reward misspecification, or distribution shift between training and evaluation.

\paragraph{Scope of evaluation.}
Finally, our evaluation focuses on scalar regression tasks and predictive distributions formed from multiple sampled rollouts. Extending the method to higher-dimensional outputs, structured scientific predictions, or settings requiring strict physical constraints remains future work. Our results should therefore be interpreted as evidence that distribution-aware RL can improve LLM-based regression in the studied settings, rather than as a replacement for well-established regression methods in all domains.

\section{Broader Impact}
\label{sec:broader_impact}

LLM-based regression has the potential to support scientific domains where prediction targets are continuous quantities and inputs are heterogeneous, such as natural-language descriptions, code, molecular strings, metadata, or experimental conditions. By improving the quality of predictive distributions, our work aims to provide a foundation for using LLMs not only as text generators, but also as flexible regression models for scientific prediction. This could help accelerate early-stage screening, experiment prioritization, and inverse-design workflows, where researchers need to reason about both expected outcomes and uncertainty before committing costly experimental or computational resources.

At the same time, these models should be used as decision-support tools rather than replacements for domain expertise or empirical validation. In scientific settings, inaccurate predictions or poorly calibrated uncertainty estimates could lead to wasted resources or misleading conclusions, especially in high-stakes areas such as materials discovery, drug development, or engineering design. Therefore, outputs from LLM-based regression systems should be interpreted cautiously, validated with domain-specific methods, and accompanied by uncertainty estimates when used for downstream decision making.

More broadly, we view this work as an initial step toward general-purpose scientific regression with language models. While the current study focuses on scalar prediction tasks, future extensions could support more complex inverse-engineering and design problems, where models propose or evaluate candidate systems under uncertainty. Responsible use will require careful benchmarking, transparent reporting of failure modes, and safeguards against over-reliance on model predictions.

\section{LLM Usage}
\label{sec:llm_usage}

LLMs are the primary model class studied in this work for regression with sampled predictive distributions. In addition, we used LLM-based tools to assist with grammar checking, wording refinement, and figure-generation code during manuscript preparation. 

\section{Task Prompts}
\label{sec:task_prompts}
We present the system prompts and user instruction prompts used for each task. For all tasks, the target value is a single real number.

\begin{tcolorbox}[
  colback=gray!10,
  colframe=gray!60,
  boxrule=0.3pt,
  arc=2pt,
  left=6pt,
  right=6pt,
  top=6pt,
  bottom=6pt
]
\textbf{Note.} The prompt boxes below show the exact system and instruction templates used in our experiments.
\end{tcolorbox}

\subsection{Synthetic Distributional Regression}

\paragraph{System Prompt.}
\begin{tcolorbox}[
  colback=gray!10,
  colframe=gray!60,
  boxrule=0.3pt,
  arc=2pt,
  breakable
]
\ttfamily
Given a synthetic regression sample, predict the target value y.\\
The input x is generated from a synthetic benchmark.\\
Output your estimate in \textbackslash boxed\{\}. Only place the numeric value inside the \textbackslash boxed\{\}.
\end{tcolorbox}

\paragraph{Instruction Prompt Template.}
\begin{tcolorbox}[
  colback=gray!10,
  colframe=gray!60,
  boxrule=0.3pt,
  arc=2pt,
  breakable
]
\ttfamily
Synthetic regression task.\\
Input feature:\\
x = \{x\}\\
\\
Predict the target value y.
\end{tcolorbox}

\paragraph{Example.}
\begin{tcolorbox}[
  colback=gray!10,
  colframe=gray!60,
  boxrule=0.3pt,
  arc=2pt,
  breakable
]
\ttfamily
Synthetic regression task.\\
Input feature:\\
x = 1.23456789\\
\\
Predict the target value y.
\end{tcolorbox}

\subsection{Coding Performance Prediction}

\paragraph{APPS System Prompt.}
\begin{tcolorbox}[
  colback=gray!10,
  colframe=gray!60,
  boxrule=0.3pt,
  arc=2pt,
  breakable
]
\ttfamily
Given an APPS sample, predict the peak memory usage in bytes.\\
Output your estimate in \textbackslash boxed\{\}. Only place the numeric value inside the \textbackslash boxed\{\}.
\end{tcolorbox}

\paragraph{KBSS System Prompt.}
\begin{tcolorbox}[
  colback=gray!10,
  colframe=gray!60,
  boxrule=0.3pt,
  arc=2pt,
  breakable
]
\ttfamily
Given a KBSS sample, predict the kernel execution latency in milliseconds.\\
Output your estimate in \textbackslash boxed\{\}. Only place the numeric value inside the \textbackslash boxed\{\}.
\end{tcolorbox}

\paragraph{Instruction Prompt Template.}
The instruction prompt is taken directly from the \texttt{input} field:
\begin{tcolorbox}[
  colback=gray!10,
  colframe=gray!60,
  boxrule=0.3pt,
  arc=2pt,
  breakable
]
\ttfamily
\{code/problem context from the dataset input field\}
\end{tcolorbox}

\paragraph{Example Structure.}
\begin{tcolorbox}[
  colback=gray!10,
  colframe=gray!60,
  boxrule=0.3pt,
  arc=2pt,
  breakable
]
\ttfamily
[Problem description, code snippet, or execution context from the dataset]
\end{tcolorbox}

\subsection{MoleculeNet Property Prediction}

\paragraph{ESOL System Prompt.}
\begin{tcolorbox}[
  colback=gray!10,
  colframe=gray!60,
  boxrule=0.3pt,
  arc=2pt,
  breakable
]
\ttfamily
Given a SMILES string, predict the aqueous solubility (logS).\\
Output your estimate in \textbackslash boxed\{\}. Only place the numeric value inside the \textbackslash boxed\{\}.
\end{tcolorbox}

\paragraph{FreeSolv System Prompt.}
\begin{tcolorbox}[
  colback=gray!10,
  colframe=gray!60,
  boxrule=0.3pt,
  arc=2pt,
  breakable
]
\ttfamily
Given a SMILES string, predict the hydration free energy in kcal/mol.\\
Output your estimate in \textbackslash boxed\{\}. Only place the numeric value inside the \textbackslash boxed\{\}.
\end{tcolorbox}

\paragraph{Lipo System Prompt.}
\begin{tcolorbox}[
  colback=gray!10,
  colframe=gray!60,
  boxrule=0.3pt,
  arc=2pt,
  breakable
]
\ttfamily
Given a SMILES string, predict the octanol/water partition coefficient.\\
Output your estimate in \textbackslash boxed\{\}. Only place the numeric value inside the \textbackslash boxed\{\}.
\end{tcolorbox}

\paragraph{Instruction Prompt Template.}
The user prompt wraps the input molecule as a SMILES string:
\begin{tcolorbox}[
  colback=gray!10,
  colframe=gray!60,
  boxrule=0.3pt,
  arc=2pt,
  breakable
]
\ttfamily
<SMILES> \{smiles\_string\} </SMILES>
\end{tcolorbox}

\paragraph{Example.}
\begin{tcolorbox}[
  colback=gray!10,
  colframe=gray!60,
  boxrule=0.3pt,
  arc=2pt,
  breakable
]
\ttfamily
<SMILES> CC(=O)OC1=CC=CC=C1C(=O)O </SMILES>
\end{tcolorbox}

\end{document}